\documentclass[11pt,twoside,twocolumn,a4paper]{article}

\usepackage{cvww}
\usepackage{times}
\usepackage{epsfig}
\usepackage{graphicx}
\usepackage{amsmath}
\usepackage{amssymb}
\usepackage{soul}
\usepackage{setspace}

\usepackage{multirow}
\usepackage{subfig}
\usepackage{tabu}
\usepackage{siunitx}
\usepackage{mathtools}
\usepackage{scalerel}
\usepackage{slashbox}
\usepackage{siunitx}
\usepackage{subcaption}
\usepackage{enumitem}

\newcommand*{\x}{\mathsf{x}\mskip1mu}
\newcommand*{\eq}{\mathsf{=}\mskip1mu}
\newlength{\imageheight}
\graphicspath{{../jpg/}{../png/}{figures/}}

\usepackage[pagebackref=true,breaklinks=true,bookmarks=false]{hyperref}
\cvwwfinalcopy

\def\cvwwPaperID{41}

\ifcvwwfinal\fi

\begin{document}
\title{EE3P: Event-based Estimation of Periodic Phenomena Properties}

\author{Jakub Kolář, Radim Špetlík, Jiří Matas\\
Visual Recognition Group, Faculty of Electrical Engineering\\
Czech Technical University in Prague\\
{\tt\small kolarj55@fel.cvut.cz}}

\newcommand{\bt}[1]{{\color{blue}{#1}}}
\newcommand{\rt}[1]{{\color{red}{#1}}}
\newcommand{\rst}[1]{{\color{red}{\st{#1}}}}

\maketitle
\ifcvwwfinal\thispagestyle{fancy}\fi


\begin{abstract}
We introduce a novel method for measuring properties of periodic phenomena with an event camera, a device asynchronously reporting brightness changes at independently operating pixels.
The approach assumes that for fast periodic phenomena,
in any spatial window where it occurs,
a very similar set of events is generated 
at the time difference corresponding to the frequency of the motion.
To estimate the frequency, we compute correlations of spatio-temporal windows in the event space.
The period is calculated from the time differences between the peaks of the correlation responses.
The method is contactless, eliminating the need for markers, and does not need distinguishable landmarks.
We evaluate the proposed method on three instances of periodic phenomena: (i) light flashes, (ii) vibration, and (iii) rotational speed. 
In all experiments, our method achieves a relative error lower than $\pm$0.04\%, which is within the error margin of ground truth measurements.
\end{abstract}

\vspace{-1em}
\section{Introduction}

The measurement of properties of periodic phenomena has a wide
applicability in diverse real-world domains.
For example,
precise quantification of rotational speed is important in many fields,  
ranging from sport analysis to 
the assessment of rotating components in machinery and mechanical systems across industries 
such as aviation (especially drones \cite{singh_drones_2018}), 
energy production using wind turbines \cite{windy_physics}, 
and motor speed testing \cite{motor_speedtesting}.

\begin{figure}
\centering
    \includegraphics[width=\linewidth]{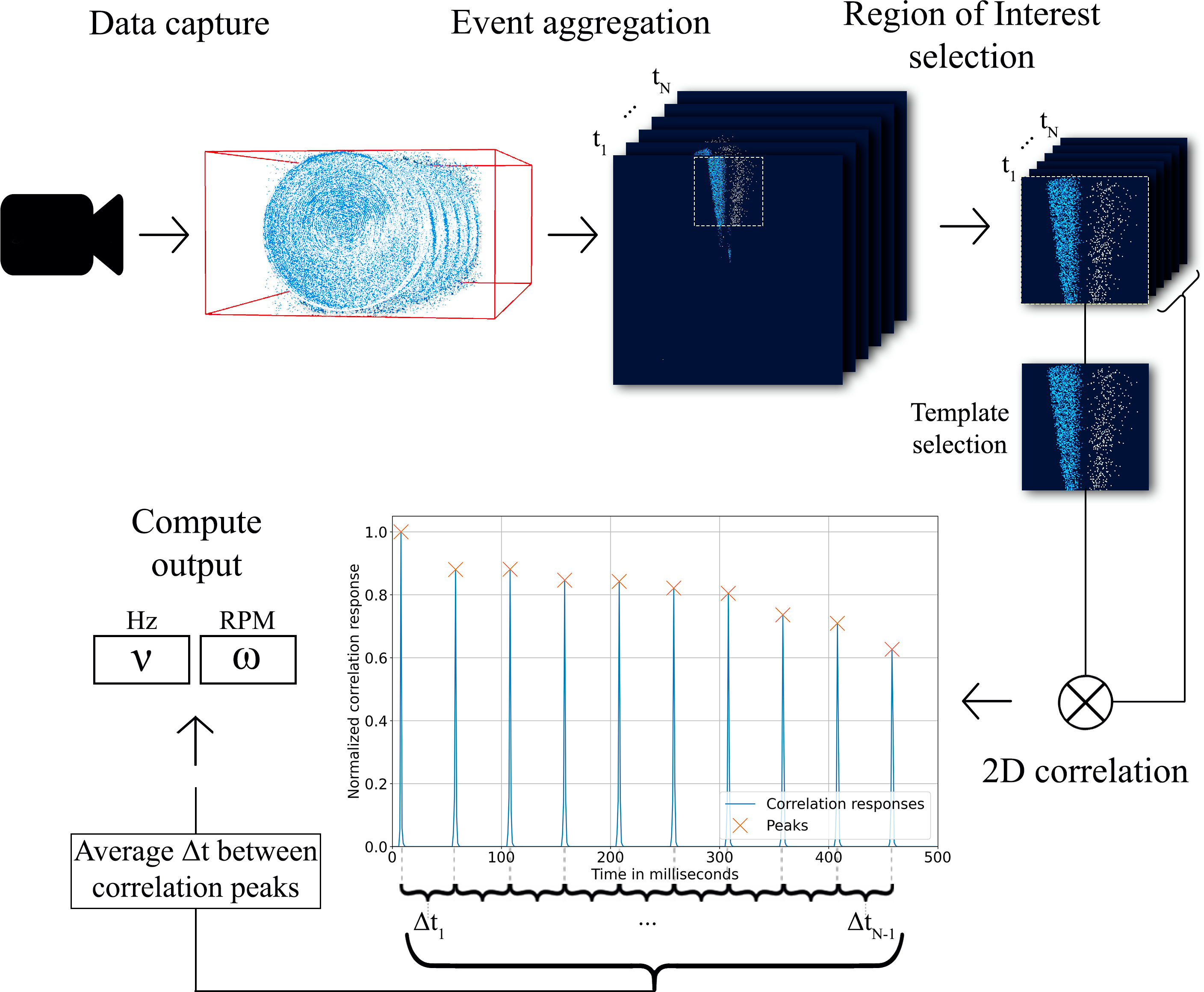}
    \caption{The proposed method: 
    (i)~data captured from an event camera is aggregated into non-overlapping arrays along the time axis,    
    (ii)~a Region of Interest and a template are selected, 
    (iii)~D correlation of the template with arrays is computed,
    (iv)~and the frequency is calculated from the average of time deltas measured between correlation peaks.}
    \label{fig:diagram}
    \vspace{-1em}
\end{figure}

Commercial devices for measuring periodic event properties, like traditional contact tachometers \cite{tachometers_2023} or rotary encoders used for measuring rotation speed, necessitate direct contact with the observed object.
These approaches interfere with the target's movement, as additional equipment must be in contact with the observed object.

In contrast to contact measurement devices, laser devices offer highly accurate \cite{ut370}, less invasive measurements. However, reflective material (\eg a sticker) must be placed on the target, reflecting the laser into the sensor while measuring. 
Under certain conditions, this limits the application of laser devices since it might not be convenient or even feasible to attach labels to particular objects or in confined spaces of the observed machinery. Another disadvantage is the device operator must aim the laser precisely at the target, as missing the reflective material pass-through results in an inaccurate measurement.

\begin{figure*}
\centering
\scriptsize
    \begin{tabular}{ccccc}
        \includegraphics[width=0.17\linewidth]{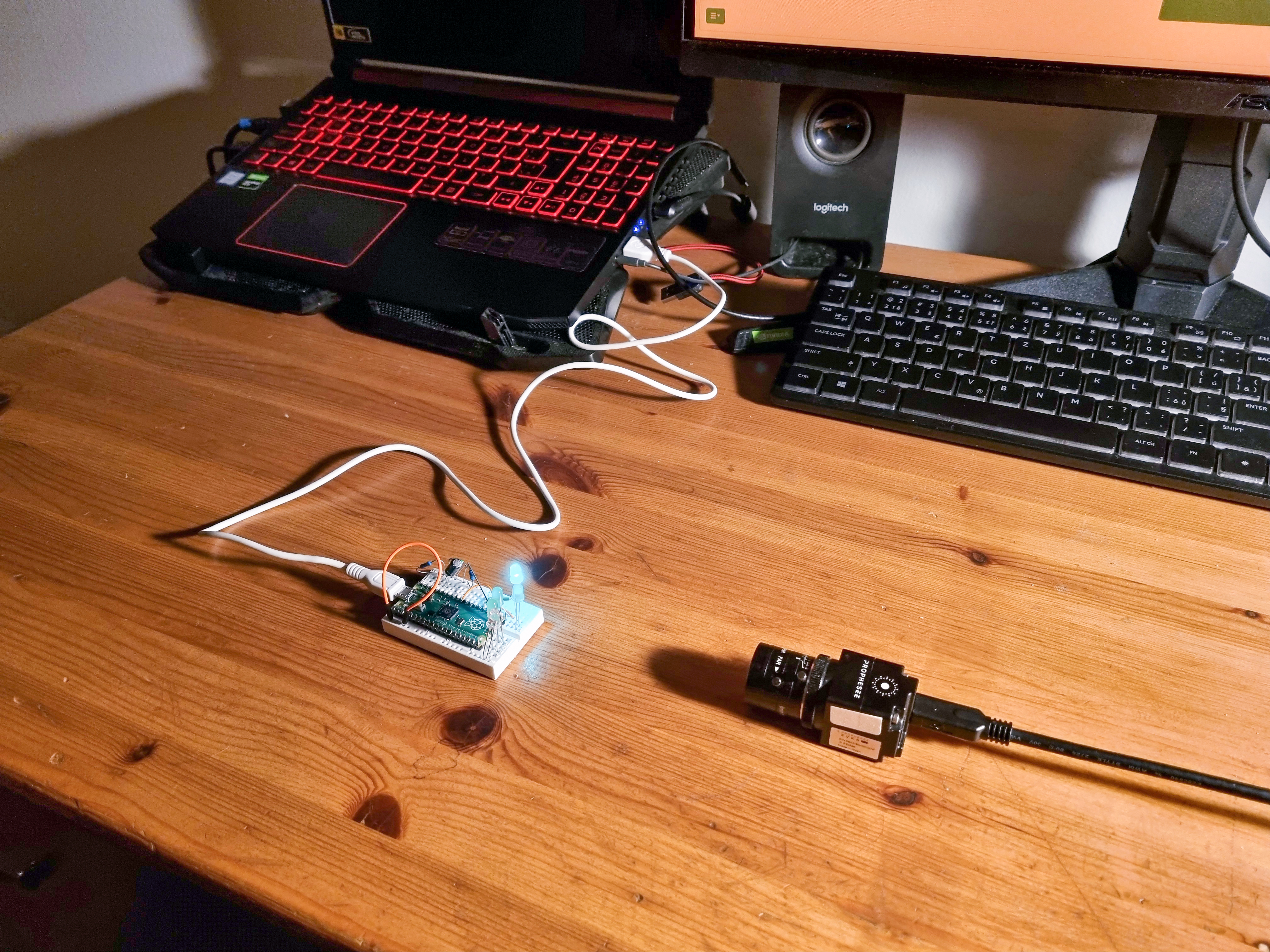} & \includegraphics[width=0.17\linewidth]{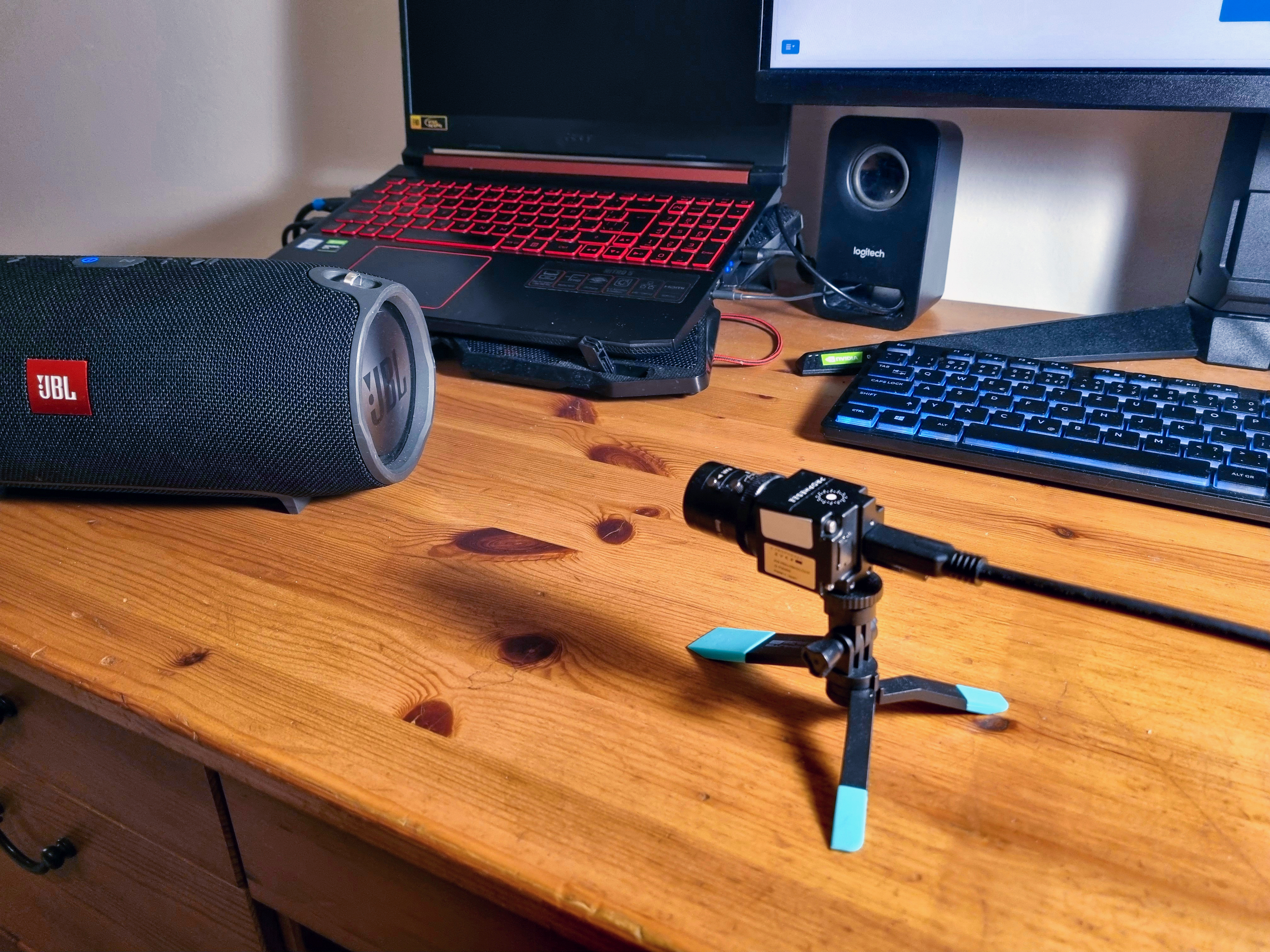} & \includegraphics[width=0.17\linewidth]{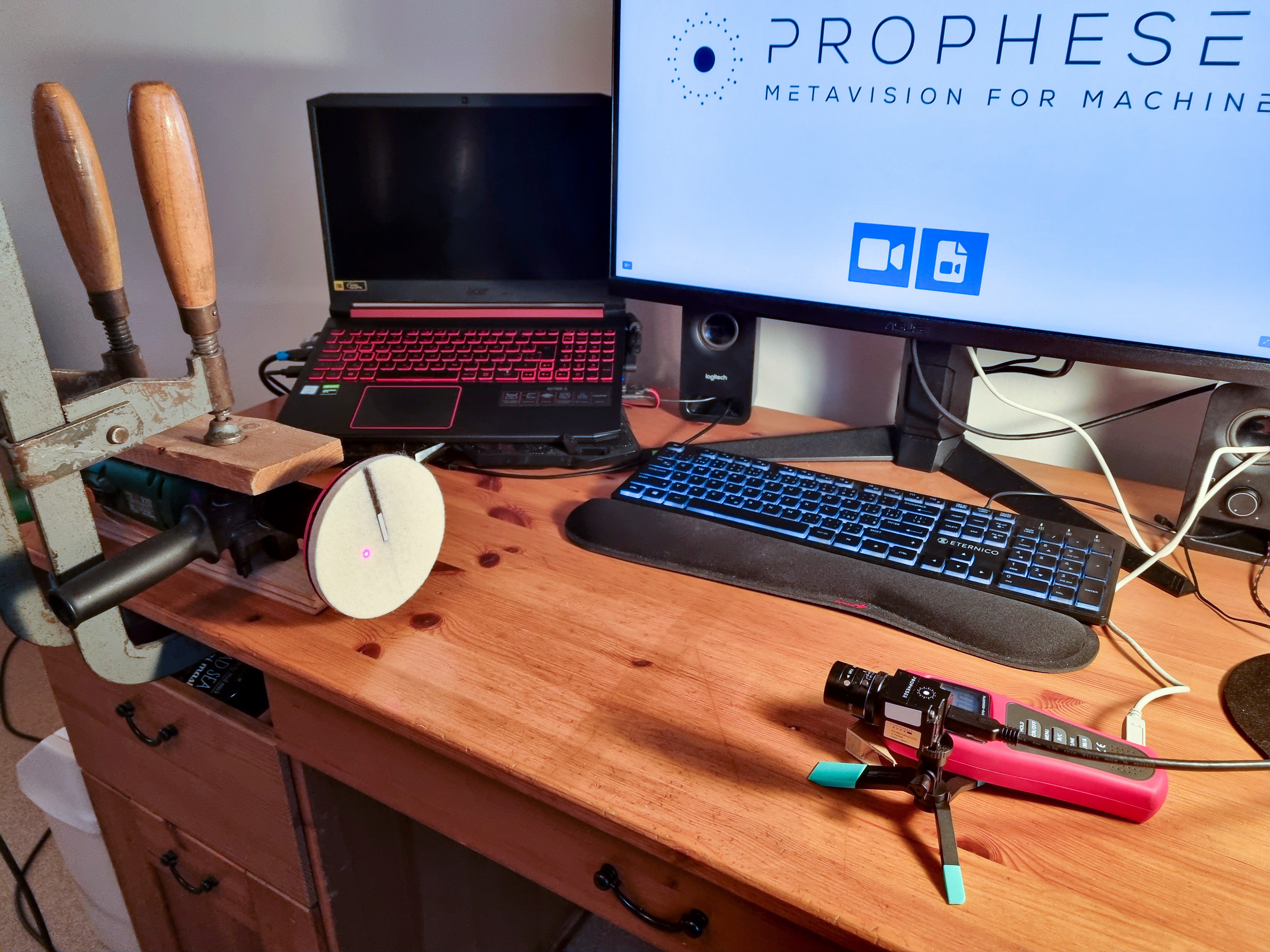} & \includegraphics[width=0.17\linewidth]{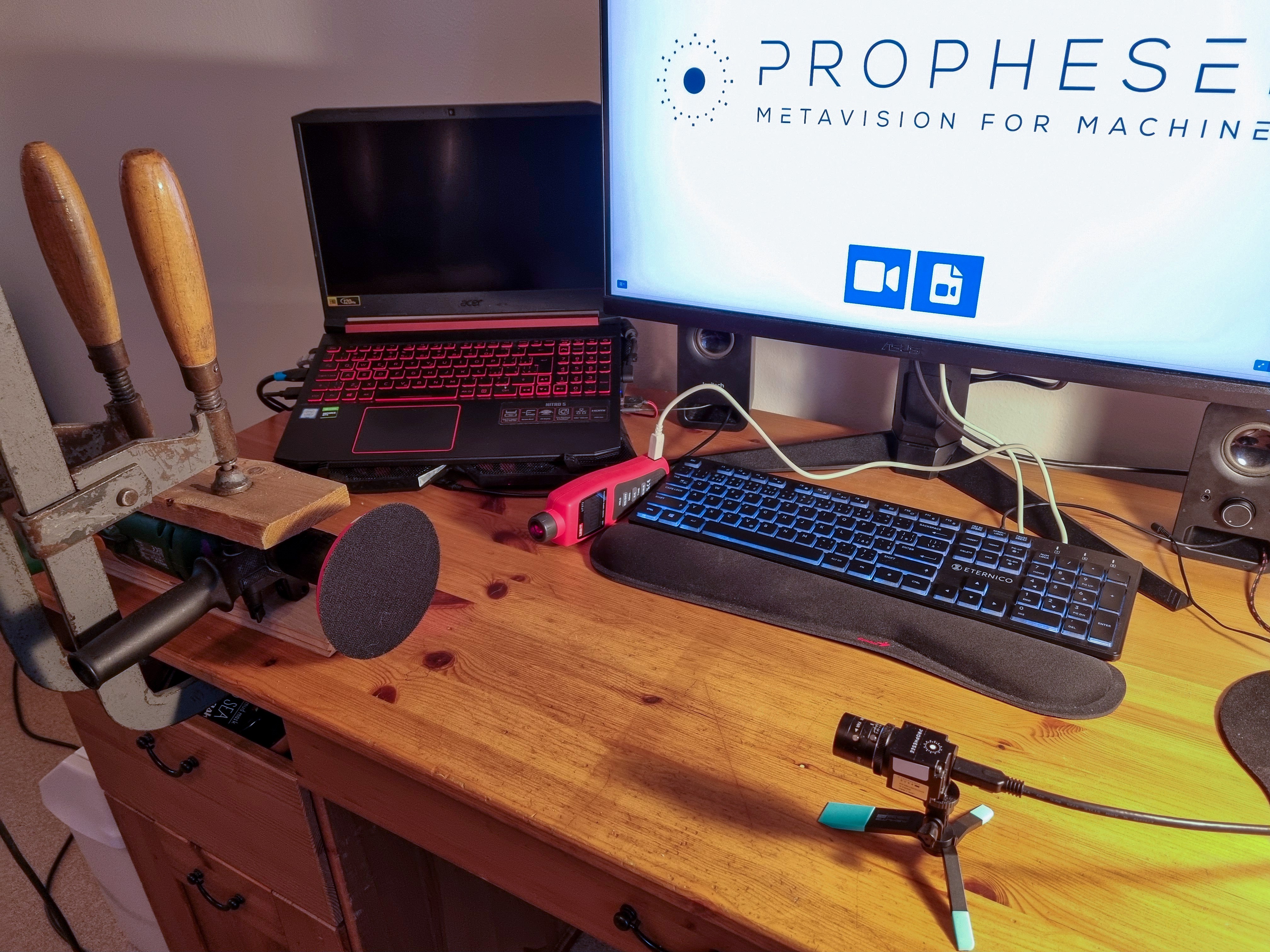} & \includegraphics[width=0.17\linewidth]{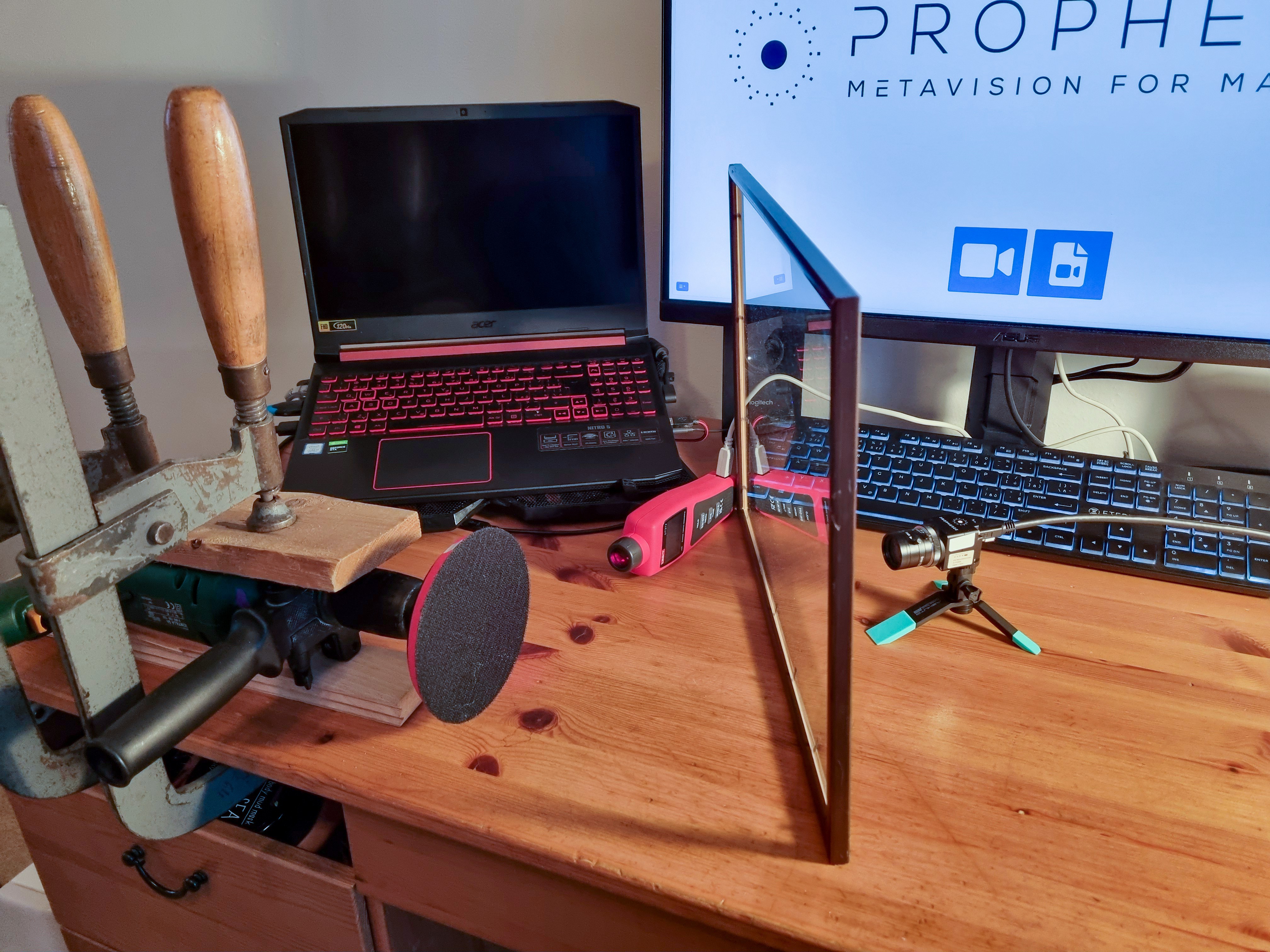}\\
        \includegraphics[width=0.17\linewidth]{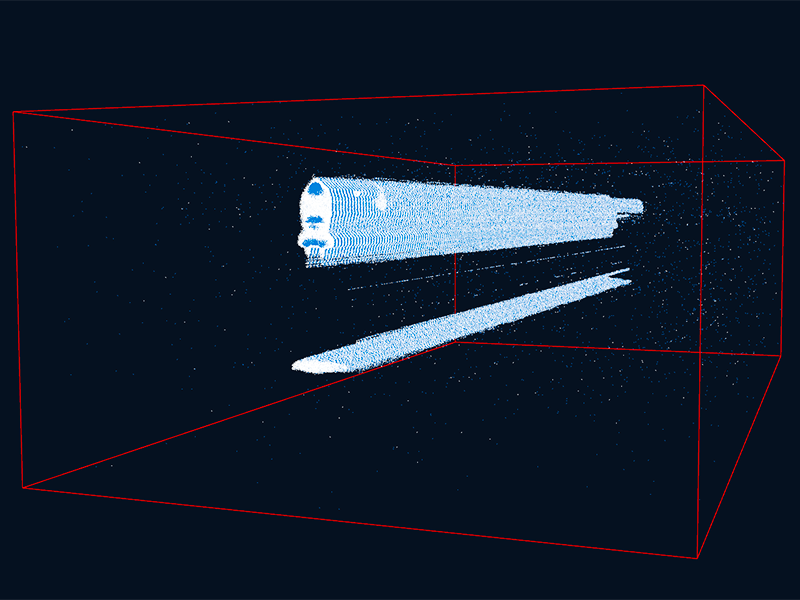} & \includegraphics[width=0.17\linewidth]{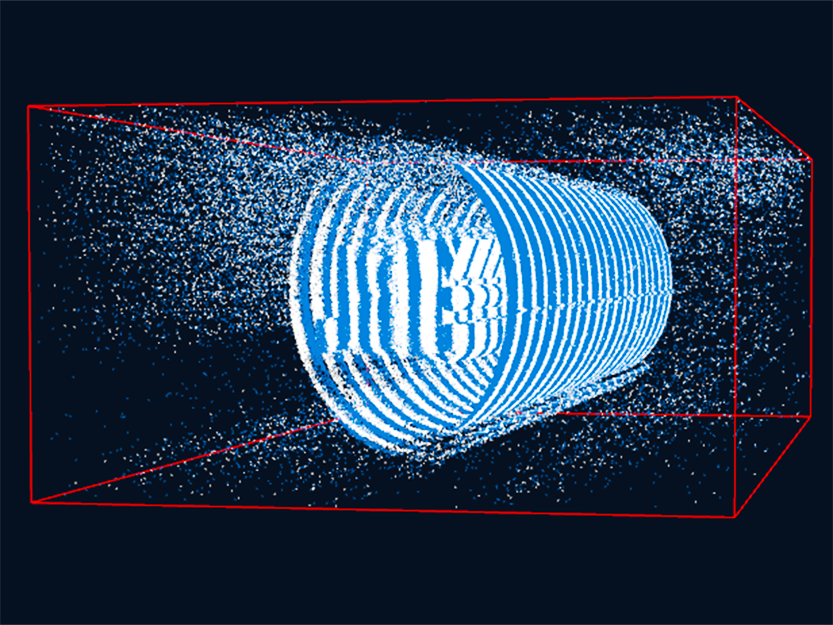} & \includegraphics[width=0.17\linewidth]{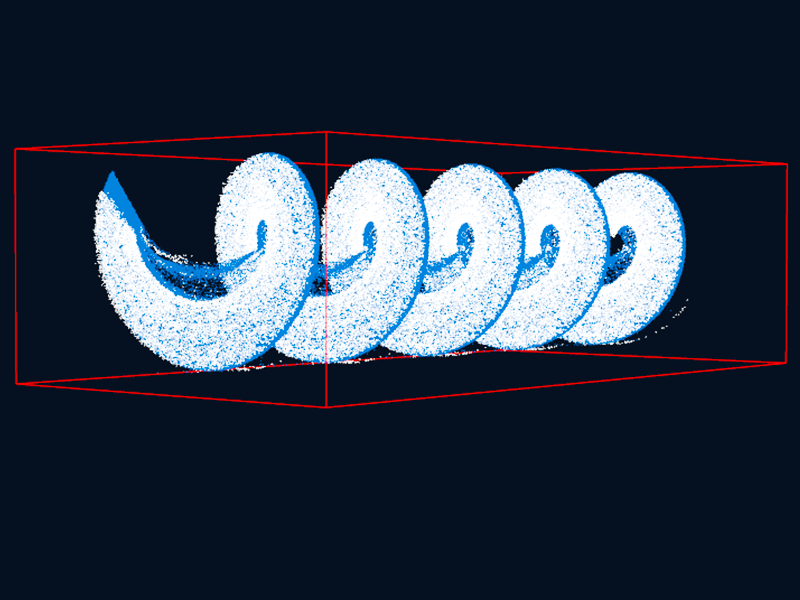} & \includegraphics[width=0.17\linewidth]{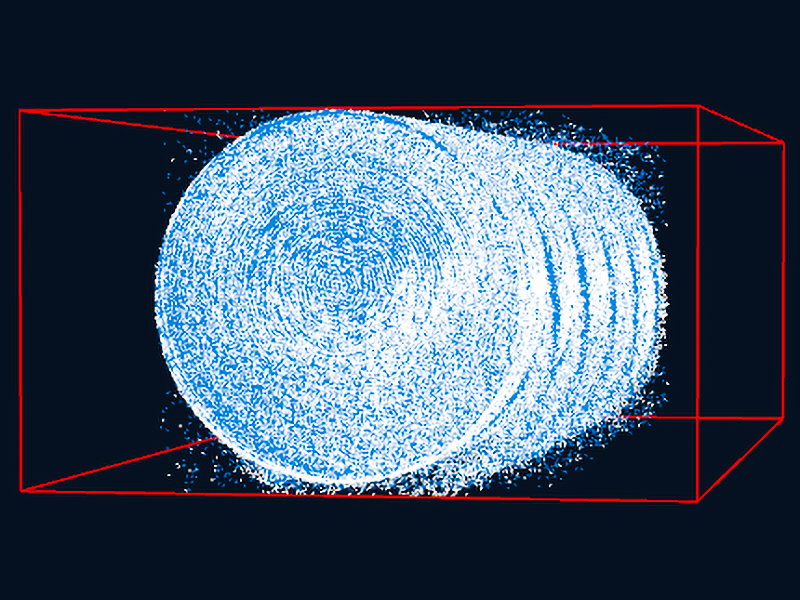} & \includegraphics[width=0.17\linewidth]{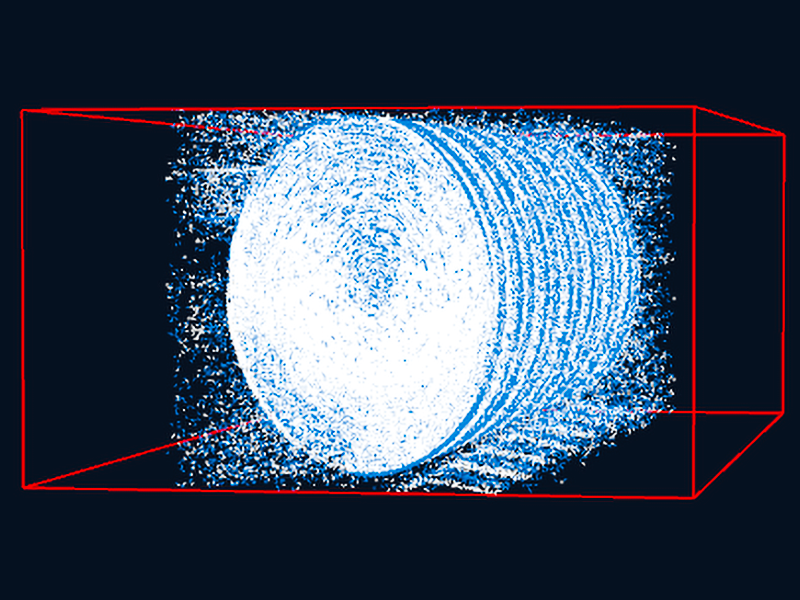}\\
        (a) flashing LED & (b) vibrating speaker & (c) a felt disc with & (d) a velcro disc, & (e) a velcro disc captured\\
        & diaphragm & a high contrast mark & fronto-parallel view & through a glass sheet\\
        & & & & at a 45\textdegree~camera angle\\
    \end{tabular}
    \caption{Experimental setup with visualisation of event camera output. Top: physical setups, bottom: events from a 250-millisecond window visualised in spatio-temporal space.
    }
    \vspace{-1em}
    \label{fig:physical_setups}
\end{figure*}

We propose a method that allows for non-contact measurement of properties of any periodic phenomena.
The proposed method computes correlations of spatio-temporal windows in the event space, 
\newline
assuming that the period of the periodic motion corresponds to the time differences between the peaks of the correlation responses (see Fig.~\ref{fig:diagram}).
The method is validated on experiments with periodic phenomena, \ie flashing light and vibration, and periodic motion, \ie rotation.
Our method achieves accuracy with a relative error of $\pm0.04\%$ in all our experiments, which falls within the margins of error of the ground truth.

\section{Related work}

In this section, we discuss existing approaches and technologies in the domain of rotation speed measurement, as periodic motion is arguably the most common periodic phenomenon. 

Firstly, we describe commercially available rotation speed measuring devices with contact and contactless options. 

Subsequently, we mention selected camera-based rotation speed measurement methods using RGB cameras, followed by event camera methods.

\subsection{Commercially Available Rotation Speed Measuring Devices}

Commercially available devices offer either contact, \eg traditional mechanical tachometers, or contact-less rotation speed measuring, \eg electrostatic and optical encoder tachometers, including laser tachometers.

Mechanical tachometers are physically attached to the target's shaft and rotate with it to determine the rotation speed. 
However, this direct physical connection introduces inaccuracies due to the mass and friction of the tachometer. 
Electrostatic sensors detect changes in the electromagnetic field caused by a shaft bearing fixed on the target, estimating the rotation speed based on the frequency of these changes. 

Optical encoder tachometers utilise a photoelectric sensor to detect light passing through a disc between a light source and the sensor. 
The disc contains opaque and transparent segments that allow for the estimation of rotation speed based on the frequency of light changes detected by the sensor. 

Laser tachometers measure rotation speed by using the frequency of laser light bounces to its sensor from small and lightweight reflective labels that must be affixed to the target's surface.

\subsection{Camera-based Rotation Speed Measurement Methods}

Wang~\etal~\cite{wang_2018} created a rotational speed measurement system based on a low-cost imaging device requiring a simple marker on the target. 
The method involves pre-processing sequential images by denoising, histogram equalisation, and circle Hough transform. 
Subsequently, these processed images undergo a similarity assessment method. 
The rotational speed is calculated by applying the Chirp-Z transform to the restructured signals, and the method achieves valid measurements with a relative error of $\pm$ 1\% in the speed range of 300 to 900 revolutions per minute (RPM).

An alternative approach~\cite{wang_2017} involved the computation of structural similarity and two-dimensional correlation between consecutive frames. 
Subsequently, similarity parameters were utilised to reconstruct a continuous and periodic time-series signal.
The fast Fourier transform was then applied to determine the period of the signal, providing the maximum relative error of $\pm$ 1\% over a speed range of 0 to 700 RPM.

Camera-based rotation speed measurement methods offer the advantage of non-contact measurements, eliminating physical attachments to the rotating object and often providing a cost-effective solution. 
However, the low frame rate of standard cameras constrains the range of observable rotating objects and potentially compromises the accuracy of speed measurements, especially for high-speed rotations.

\subsection{Event-based Rotation Speed Measurement Methods}

Hylton~\etal~\cite{hylton_experiments_2021} introduced a technique for computing the optical flow of a moving object within an event stream, demonstrating its application in estimating the rotational speed of a disc with a black-and-white pattern. 
However, the algorithm's design lacked the sophistication required to deal with the non-structural and noisy event stream to obtain accurate measurements of high-speed rotation.

EV-Tach method~\cite{zhao_high_2022} starts by eliminating event outliers by estimating the median distance from events to their centroid, flagging events with distances surpassing a specified threshold as outliers. 
Subsequently, it identifies rotating objects characterised by centrosymmetric shapes and proceeds to track specific features, such as propeller blades. 

Event-based rotation speed measurement methods offer the advantage of high temporal resolution, enabling precise tracking of rapid rotational motion. 
However, these methods may face challenges in scenarios where clear observable landmarks or markers on the rotating target are absent, limiting their applicability in specific environments and necessitating well-defined visual features for accurate measurements or knowledge of the centre of rotation.

\section{Proposed method}
In this section, we introduce our method. Put simply, our method first aggregates outputs of an event camera
\footnote{
The data acquired from the event camera are represented as a list of tuples $(x, y, t, p)$, where $x$ and $y$ denote spatial coordinates of the event, $t$ the timestamp of the event, and $p$ is the polarity of the brightness change. 
The $p$ value is $-1$ in a case of brightness decrease, $1$ in a case of brightness increase, and $0$ in a case of no brightness change larger than the defined threshold. The list contains events with ascending timestamps.}along the time axis, then computes a 2D correlation of a selected template with the aggregated data in a selected region of interest and outputs the average duration between peaks of correlation responses. A detailed description follows.

The \emph{Region of Interest} (RoI) is a two-dimensional square area represented by four coordinates defining its top-left and bottom-right corners.
We aggregate events within fixed time intervals with spatial coordinates within the selected RoI into two-dimensional arrays we call \emph{event-aggregation arrays}. 

The aggregation procedure starts by creating a two-dimensional array filled with zeros of the same size as the RoI.
We go through the list of events with spatial coordinates within a chosen Region of Interest (RoI) and timestamps within a specified time interval. 
For each of these events, we modify an element in an array. 
The position of this element in the array corresponds to the spatial coordinates of the event relative to the RoI, and the polarity of the event determines the new value in the array.

\begin{figure}
\centering
    \begin{tabular}{@{}c @{\hspace{0.7em}}c @{\hspace{0.7em}}c @{\hspace{0.7em}}c@{}}
        \includegraphics[width=0.22\columnwidth]{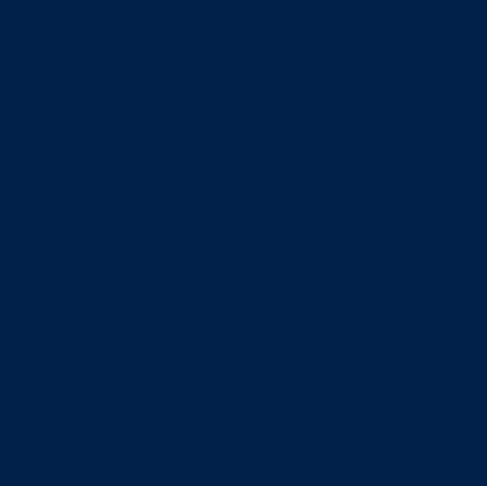} & \includegraphics[width=0.22\columnwidth]{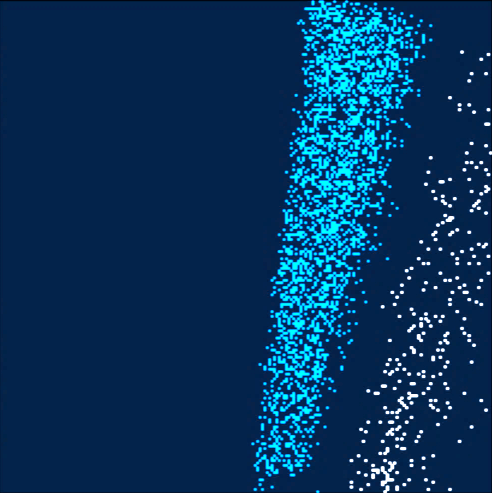} & \includegraphics[width=0.22\columnwidth]{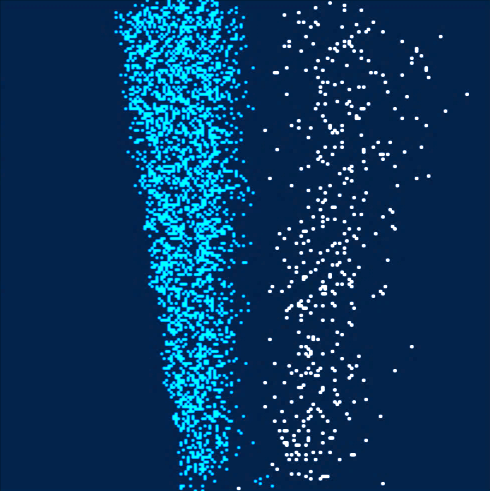} & \includegraphics[width=0.22\columnwidth]{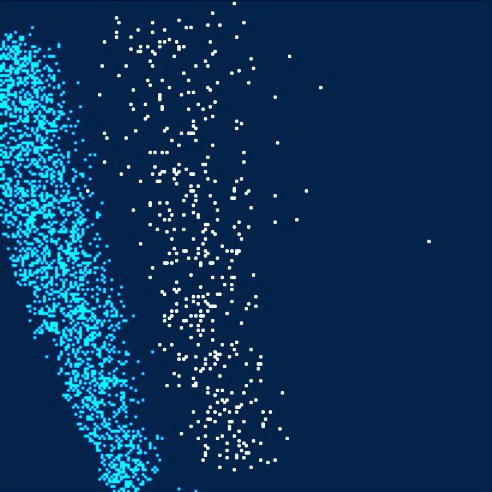} \\
        \scriptsize $t \in \lbrack50, 51)$ & \scriptsize $t \in \lbrack57, 58)$ & \scriptsize $t \in \lbrack59, 60)$ & \scriptsize $t \in \lbrack61, 62)$\\ 
    \end{tabular}
    \caption{Aggregated events in a fixed time interval of one millisecond for a selected Region of Interest. Positive events are represented by white color, and negative events are bright blue.}
    \vspace{-1.25em}
    \label{fig:line_timewindows}
\end{figure}

We choose one of these event-aggregation arrays to calculate the correlation with all event-aggregation arrays. 
We refer to this selected array as a \emph{template}.
Figure~\ref{fig:line_timewindows} visualises four selected event-aggregation arrays, showcasing the spatial distribution of events within one millisecond set time intervals.

Next, we calculate correlation responses between the template and all event-aggregation arrays. 

As expected, the responses have periodic peaks. When a peak is reached, it signifies the completion of one period of the phenomenon. 

An example of periodic peaks in the correlation responses is shown in Fig.~\ref{fig:diagram}. 

Subsequently, we compute $N-1$ delta times $\Delta t_{i}$ by measuring the temporal differences between successive $N$ event-aggregation arrays that exhibit peaks in their correlation responses. Each $\Delta t_{i}$ represents the microseconds it takes for the observed object to complete one revolution or cycle of states.
\newpage
The RPM value based on a single revolution is subsequently calculated using the following formula for experiments on measuring the speed of rotating objects
\begin{equation}
    \mbox{RPM}_{i} = \frac{10^{6}}{\Delta t_{i}} \times 60, i = 1,2,...,N-1
\end{equation}
where $N$ is the number of event-aggregation arrays.
Ultimately, we calculate the average RPM value for each second of data as
\begin{equation}
    \overline{\mbox{RPM}} = \frac{\sum_{i=1}^{M} \mbox{RPM}_i}{M}
\end{equation}
where $M$ is the number of samples in one second of data.

For the other experiments (\ref{experiment:led}, \ref{experiment:speaker}) the frequency $\nu$ expressed in hertz (Hz) is computed for each $\Delta t_{i}$ as
\begin{equation}
    \nu_{i} = \frac{10^{6}}{\Delta t_{i}}, i = 1,2,...,N-1
\end{equation}
We then calculate the arithmetic mean frequency of periodic movement.

The $\sigma$~value in Tab.~\ref{tab:roisizes_line}-\ref{tab:windowdur_velcroside} is computed as
\begin{equation}
\label{eq:sigma}
    \sigma = \sqrt{\frac{\sigma^{2}}{M}}
\end{equation} 
where $M$ is the measurement count during the respective time interval of $1$ second. The $\sigma$ value represents the standard deviation of the average measured value. We assume the measurements are independently identically distributed and drawn from a normal distribution. We chose the confidence interval of $95.4\%$, by which our point estimate of the mean should be less than $2\sigma$ away from the true mean, and our point estimate of standard deviation should be less than $2\sigma$.

Our proposed method requires parameters that need to be selected by the user. These parameters are the event aggregation duration, position and size of the RoI and which event-aggregation array to use as a template for calculating correlation responses. An overview of the method can be seen in Fig.~\ref{fig:diagram}.

\section{Experiments}

This section presents five experiments demonstrating the versatility of our method for frequency and rotational speed estimation. The first two experiments focus on frequency measurement. Experiment \ref{experiment:led} measures the frequency of a flashing diode, while Experiment \ref{experiment:speaker} estimates the frequency of vibration.

The following three experiments investigate rotational speed measurement. Experiment \ref{experiment:felt-disc} measures the speed of a high-contrast disc, while Experiment \ref{experiment:velcro-disc-front} measures a disc with uniform velcro, where identifying patterns is challenging. In both these experiments, the event camera maintains a fronto-parallel position relative to the disc.

Experiment \ref{experiment:velcro-disc-side} showcases the method's robustness. It demonstrates accurate measurements even when the camera axis is not aligned with the rotation axis, resulting in elliptical object trajectories on the 2D image plane. Furthermore, the experiment confirms that capturing data through transparent materials does not compromise accuracy.

For physical setups of experiments, see Fig.~\ref{fig:physical_setups}.
We maintain the same lighting conditions and stationary position of sensors and the observed object.
\subsection{Sensor setup}
An overview of both the event camera and the laser tachometer follows.

\vspace{-1em}\paragraph{Event camera}In our experiments, we used the Prophesee~EVK4~HD event camera, a device asynchronously reporting brightness changes (events) at independently operating pixels. 
The camera's resolution is $1280\x720$ pixels and can capture up to 1066 million events per second \cite{ev_cam_specs}.

The event sensor's behavior can be fine-tuned using five adjustable biases \cite{evk_biases}:
\begin{itemize}[leftmargin=1em]
    \item \textbf{Contrast sensitivity threshold biases:} These biases control the sensor's sensitivity to changes in light intensity. \verb|diff_on| determines the minimum amount of brightness increase required to trigger a positive event, while \verb|diff_off| sets the threshold for a negative event to be caused by a decrease in brightness.
    \item \textbf{Bandwidth biases:} These biases control filtering applied to the incoming light signal. \verb|fo| adjusts a low-pass filter that removes rapid light fluctuations, while \verb|hpf| controls a high-pass filter that eliminates slow changes in illumination.
    \item \textbf{Dead time bias:} This bias, \verb|refr|, sets the duration for which a pixel remains inactive after registering an event.
\end{itemize}

In our experiments, we set the contrast sensitivity threshold biases to 20 and the high-pass filter bias to 50. Other biases remained at their default values.

\vspace{-1em}\paragraph{Laser tachometer} To capture the ground truth (GT) rotation speed data, the Uni-Trend UT372 laser tachometer \cite{ut370} was used. The tachometer range is 10 to 99~999 RPM with a relative error of $\pm0.04\%$.
We chose the lowest available measurement output rate of $0.5$ seconds and captured the measurements via a USB cable.
It is worth mentioning that the optical tachometer outputs only $3$ to $5$ samples per second, while our method produces a measurement for each period of the observed periodic phenomenon.

The GT frequency is known for the other experiments as it was manually set beforehand.

In the following subsections, we present the mentioned experiments and the results of our method. In each subsection, the selection of RoI and the duration of the event aggregation are discussed, as we hypothesised that these two parameters influence our method the most.

\subsection{Measuring periodic light flashes}
\label{experiment:led}

\begin{table}
\vspace{0.5em}
\begin{tabular}{p{0.95\columnwidth}}
\begin{minipage}{\linewidth}

    \centering
    \scriptsize
    \setlength\tabcolsep{0.5pt}
    \begin{tabular}{c}
    \settoheight{\imageheight}{\includegraphics{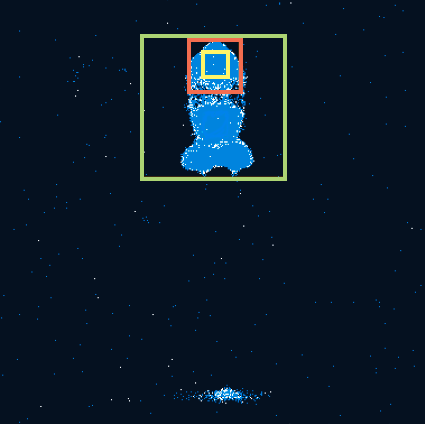}}
    \includegraphics[trim=0 0.65\imageheight{} 0 0.05\imageheight{}, clip, width=0.95\columnwidth]{led_rois_findsize+}\\
    (a) Selected Regions of Interest\\
    \end{tabular}

    \centering
    \scriptsize
    \setlength\tabcolsep{2.5pt}
    \begin{tabular}{ccc}
    \includegraphics[width=0.3\columnwidth]{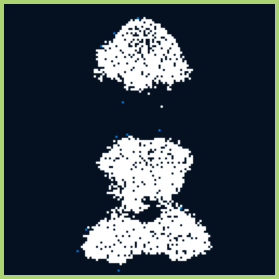} & \includegraphics[width=0.3\columnwidth]{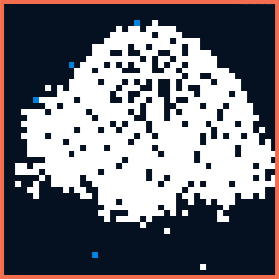} & \includegraphics[width=0.3\columnwidth]{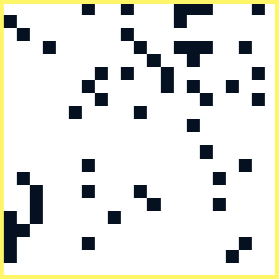}\\
    $125\x125$ px & $45\x45$ px & $20\x20$ px\\
    \multicolumn{3}{c}{\begin{tabular}[c]{@{}c@{}}(b) Templates with aggregation duration of 0.1 ms.\end{tabular}}\\
    \end{tabular}

    \centering
    \scriptsize
    \setlength\tabcolsep{1.6pt}
    \begin{tabular}{cccc}
    \includegraphics[width=0.225\columnwidth]{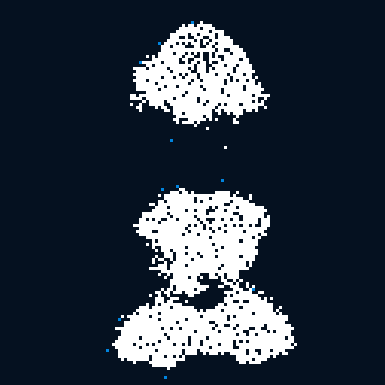} &  \includegraphics[width=0.225\columnwidth]{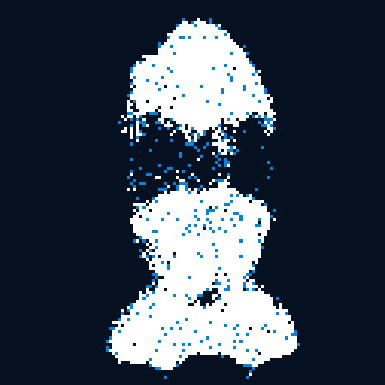} & \includegraphics[width=0.225\columnwidth]{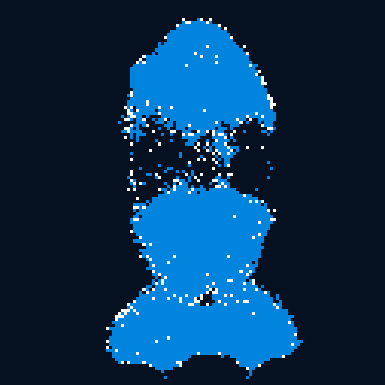} &  \includegraphics[width=0.225\columnwidth]{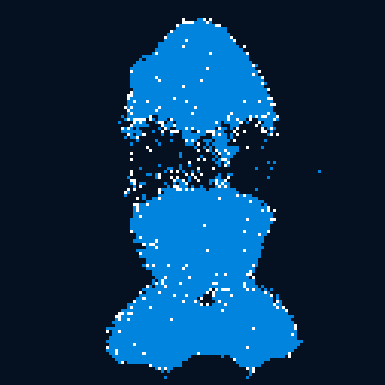}\\
    $t\eq0.1$ ms & $t\eq0.25$ ms & $t\eq0.5$ & $t\eq1$ ms\\
    \multicolumn{4}{c}{\begin{tabular}[c]{@{}c@{}}(c) A template as a function of\end{tabular}}\\
    \multicolumn{4}{c}{\begin{tabular}[c]{@{}c@{}}the duration $t$ of the aggregation time interval.\end{tabular}}
    \end{tabular}

\vspace{-1.0em}\captionof{figure}{Setup of experiment~\ref{experiment:led} (see Tab.~\ref{tab:roisizes_led}, \ref{tab:windowdur_led}).}
\vspace{1.0em}
\label{fig:led_pics}

    \centering
    \tiny
    \newcolumntype{C}{>{\centering\arraybackslash}p{1.3cm}}
    \begin{tabular}{|C||C|C|C|C|}
        \hline
    \backslashbox{\,t(s)}{method\,} & \multicolumn{1}{c|}{\begin{tabular}[c]{@{}c@{}}\,Ground\,\\ truth\end{tabular}} & 
    \multicolumn{1}{c|}{\begin{tabular}[c]{@{}c@{}}Our method\\$125\x125$px\end{tabular}} & 
    \multicolumn{1}{c|}{\begin{tabular}[c]{@{}c@{}}Our method\\$45\x45$px\end{tabular}} & 
    \begin{tabular}[c]{@{}c@{}}Our method\\ $20\x20$px\end{tabular} \\ \hline\hline
         $\lbrack0,4)$ & \begin{tabular}[c]{@{}c@{}}2000\end{tabular} & \begin{tabular}[c|]{@{}c@{}}2000\\$\pm$ 0\end{tabular} & 
                    \begin{tabular}[c]{@{}c@{}}2000\\$\pm$ 0\end{tabular} & \begin{tabular}[c]{@{}c@{}}2000.54\\$\pm$ 0.82\end{tabular}\\ \hline
    \end{tabular}
    \vspace{-1.0em}
    \captionof{table}{Frequency (Hz) $\pm$ 2$\sigma$~(\ref{eq:sigma}) as a function of the size of the Region of Interest (see Fig.~\ref{fig:physical_setups}a,~\ref{fig:led_pics}b).}
    \vspace{1.5em}
    \label{tab:roisizes_led}

    \centering
    \tiny
    \newcolumntype{C}{>{\centering\arraybackslash}p{2.25cm}}
    \begin{tabular}{|C||C|C|}
        \hline  
        {\backslashbox{\,t(ms)}{method\,}}& Ground truth & 
        \multicolumn{1}{c|}{\begin{tabular}[c]{@{}c@{}}Our method\\$60\x60$px\end{tabular}}\\ \hline\hline
        0.1 & \multirow{4}{*}{\begin{tabular}[c]{@{}c@{}}2000\end{tabular}} & \begin{tabular}[c]{@{}c@{}}2000\\$\pm$0\end{tabular} \\ \cline{1-1} \cline{3-3} 
        0.25 &  & \begin{tabular}[c]{@{}c@{}}2000\\$\pm$ 0\end{tabular} \\ \cline{1-1} \cline{3-3} 
        0.5 &  & \begin{tabular}[c]{@{}c@{}}757.79\\$\pm$ 17.28\end{tabular} \\ \cline{1-1} \cline{3-3} 
        1.0 &  & \begin{tabular}[c]{@{}c@{}}361.81\\$\pm$ 12.74\end{tabular} \\ \hline
    \end{tabular}
    \vspace{-1.0em}
    \captionof{table}{Frequency (Hz) $\pm$ 2$\sigma$~(\ref{eq:sigma}) as a function of the aggregation time interval (see Fig.~\ref{fig:physical_setups}a,~\ref{fig:led_pics}c).}
    \label{tab:windowdur_led}
    \vspace{-2.25em}
\end{minipage}
\end{tabular}
\end{table}

In this experiment, we used a simple circuit with a diode (see Fig.~\ref{fig:physical_setups}a) and a software-defined oscilloscope based on the Raspberry Pi Pico controller \cite{fiala_raspberry_2023} to precisely set the flashing frequency and portion of the period (duty cycle) that the diode should emit light. We opted for 2000~Hz and a duty cycle of 50\%.

\vspace{-1em}\paragraph{Selection of RoI}
We selected three distinct Regions of Interest (RoIs) with varying positions and sizes. The first RoI covers the entirety of the flashing diode, while the second focuses solely on its upper half. Lastly, the third RoI is set to a smaller area within the upper portion of the diode. 
The results are in Tab.~\ref{tab:roisizes_led}.
Our method consistently estimates the flashing frequency with minimal accuracy decrease, even for the smallest RoI, demonstrating robustness and flexibility in RoI selection.

\vspace{-1em}\paragraph{Selection of aggregation duration}

We fixed the $125\x125$ pixel RoI mentioned in the preceding paragraph and conducted experiments by adjusting the duration of the event-aggregation window. 
Our method consistently achieves precise measurements for durations up to 0.25 milliseconds, as can be seen in Tab.~\ref{tab:windowdur_led}. However, exceeding this significantly degrades accuracy due to the aggregation of events from multiple flashes within a single event-aggregation array. This highlights the importance of selecting an appropriate aggregation duration based on the expected frequency range to ensure reliable measurements.

\subsection{Measuring vibrations}
\label{experiment:speaker}

\begin{table}
\vspace{0.5em}
\begin{tabular}{p{0.95\columnwidth}}
\begin{minipage}{\linewidth}

    \centering
    \scriptsize
    \setlength\tabcolsep{0pt}
    \begin{tabular}{c}
    \settoheight{\imageheight}{\includegraphics{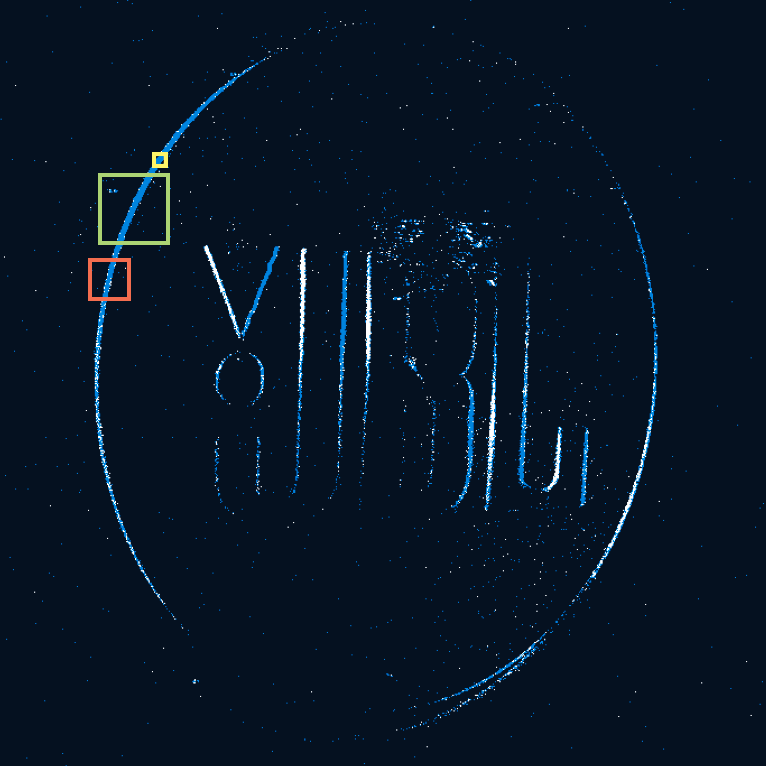}}
    \includegraphics[trim=0 0.55\imageheight{} 0 0.15\imageheight{}, clip, width=0.95\columnwidth]{speaker_rois_findsize+}\\
    (a) Selected Regions of Interest\\
    \end{tabular}

    \centering
    \scriptsize
    \setlength\tabcolsep{2.5pt}
    \begin{tabular}{ccc}
    \includegraphics[width=0.3\columnwidth]{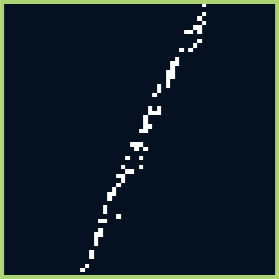} & \includegraphics[width=0.3\columnwidth]{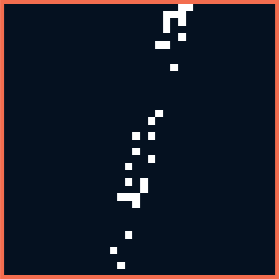} & \includegraphics[width=0.305\columnwidth]{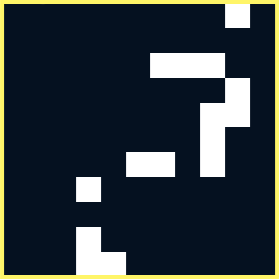}\\
    $60\x60$ px & $35\x35$ px & $10\x10$ px\\
    \multicolumn{3}{c}{\begin{tabular}[c]{@{}c@{}}(b) Templates with aggregation duration of 0.25 ms.\end{tabular}}\\
    \end{tabular}

    \centering
    \scriptsize
    \setlength\tabcolsep{1.35pt}
    \begin{tabular}{cccc}
    \includegraphics[width=0.225\columnwidth]{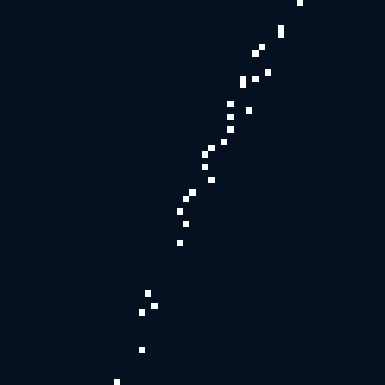} &  \includegraphics[width=0.225\columnwidth]{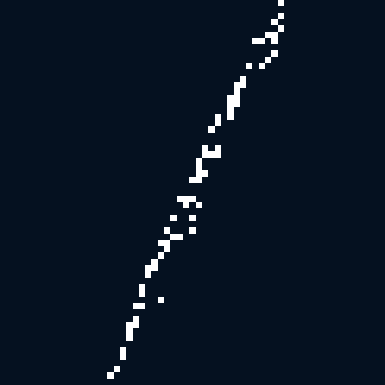} & \includegraphics[width=0.225\columnwidth]{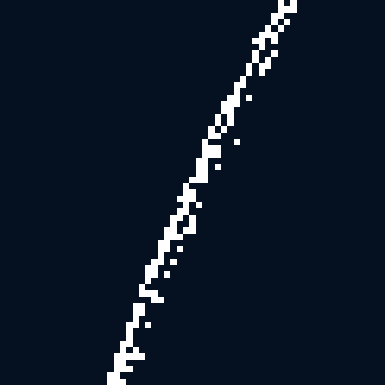} &  \includegraphics[width=0.225\columnwidth]{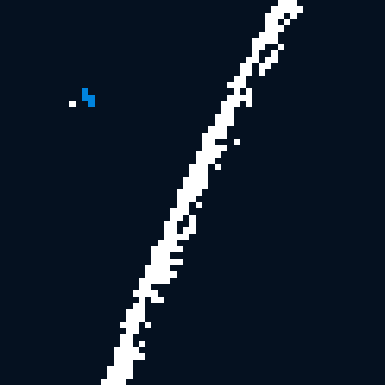}\\
    $t\eq0.1$ ms & $t\eq0.25$ ms & $t\eq0.5$ & $t\eq1$ ms\\
    \multicolumn{4}{c}{\begin{tabular}[c]{@{}c@{}}(c) A template as a function of\end{tabular}}\\
    \multicolumn{4}{c}{\begin{tabular}[c]{@{}c@{}}the duration $t$ of the aggregation time interval.\end{tabular}}
    \end{tabular}

\vspace{-1.0em}
\captionof{figure}{Setup of experiment~\ref{experiment:speaker} (see Tab.~\ref{tab:roisizes_speaker}, \ref{tab:windowdur_speaker}).}
\vspace{2em}
\label{fig:speaker_pics}

    \centering
    \tiny
    \newcolumntype{C}{>{\centering\arraybackslash}p{1.3cm}}
    \begin{tabular}{|C||C|C|C|C|}
        \hline
    \backslashbox{\,t(s)}{method\,} & \multicolumn{1}{c|}{\begin{tabular}[c]{@{}c@{}}\,Ground\,\\ truth\end{tabular}} & 
    \multicolumn{1}{c|}{\begin{tabular}[c]{@{}c@{}}Our method\\$60\x60$px\end{tabular}} & 
    \multicolumn{1}{c|}{\begin{tabular}[c]{@{}c@{}}Our method\\ $35\x35$px\end{tabular}} & 
    \begin{tabular}[c]{@{}c@{}}Our method\\ $10\x10$px\end{tabular} \\ \hline\hline
         $\lbrack0,4)$ & \begin{tabular}[c]{@{}c@{}}98\end{tabular} & \begin{tabular}[c|]{@{}c@{}}98.1\\$\pm$ 0.63\end{tabular} & 
                    \begin{tabular}[c]{@{}c@{}}98.21\\$\pm$ 0.73\end{tabular} & \begin{tabular}[c]{@{}c@{}}98.48\\$\pm$ 0.9\end{tabular}\\ \hline
    \end{tabular}
    \vspace{-1.0em}
    \captionof{table}{Frequency (Hz) $\pm$ 2$\sigma$~(\ref{eq:sigma}) as a function of the size of the Region of Interest (see Fig.~\ref{fig:physical_setups}b,~\ref{fig:speaker_pics}b).}
    \vspace{2em}
    \label{tab:roisizes_speaker}

    \centering
    \tiny
    \newcolumntype{C}{>{\centering\arraybackslash}p{2.25cm}}
    \begin{tabular}{|C||C|C|}
        \hline  
        {\backslashbox{\,t(ms)}{method\,}}& Ground truth & 
        \multicolumn{1}{c|}{\begin{tabular}[c]{@{}c@{}}Our method\\$125\x125$px\end{tabular}}\\ \hline\hline
        0.1 & \multirow{4}{*}{\begin{tabular}[c]{@{}c@{}}98\end{tabular}} & \begin{tabular}[c]{@{}c@{}}98.39\\$\pm$1.5\end{tabular} \\ \cline{1-1} \cline{3-3} 
        0.25 &  & \begin{tabular}[c]{@{}c@{}}98.15\\$\pm$ 1.13\end{tabular} \\ \cline{1-1} \cline{3-3} 
        0.5 &  & \begin{tabular}[c]{@{}c@{}}98\\$\pm$ 0.89\end{tabular} \\ \cline{1-1} \cline{3-3} 
        1.0 &  & \begin{tabular}[c]{@{}c@{}}98.05\\$\pm$ 0.88\end{tabular} \\ \hline
    \end{tabular}
    \vspace{-1.0em}
    \captionof{table}{Frequency (Hz) $\pm$ 2$\sigma$~(\ref{eq:sigma}) as a function of the aggregation time interval (see Fig.~\ref{fig:physical_setups}b,~\ref{fig:speaker_pics}c).}
    \label{tab:windowdur_speaker}
    \vspace{-2em}

\end{minipage}
\end{tabular}
\end{table}

For this experiment, we utilised a Bluetooth speaker equipped with two large diaphragms specifically designed for reproducing low frequencies. To precisely control the emitted sound, we employed an Android application \cite{luxdelux_frequency_2023} that enabled us to select and play a specific frequency. We chose 98~Hz, corresponding to the musical note $G_{2}$ within the standard tuning system where $A_{4}$ is set to 440~Hz. The event camera captured the speaker's vibrating diaphragm for four seconds (see Fig.~\ref{fig:physical_setups}b).

\vspace{-1em}\paragraph{Selection of RoI}
We experimented with different RoI positions and sizes to find the smallest RoI size that still produced accurate results.
We picked three of them for demonstration purposes (see Fig.~\ref{fig:speaker_pics}a,b), with the duration of the event aggregation set to $0.25$~millisecond. 
The results for four seconds of data, capturing 406 individual vibrations, are presented in Table~\ref{tab:roisizes_speaker}. Notably, even a tiny RoI of 10x10 pixels yields accurate results with minimal deviation from the ground truth. This highlights the method's potential for practical applications where reducing the computational load or focusing on minor specific areas of interest is benefitial.

\vspace{-1em}\paragraph{Selection of aggregation duration}
Our experiment on varying aggregation durations (Table~\ref{tab:windowdur_speaker}) reveals that the optimal aggregation duration in this scenario is 0.5 milliseconds. Interestingly, we observed a general trend of increasing accuracy with longer durations, suggesting that aggregating more events leads to a more distinct pattern in the template, potentially improving the method's ability to distinguish between periodic and non-periodic event-aggregation arrays.

\subsection{Measuring rotation speed}

This section presents three experiments investigating rotational speed measurement using a power drill to spin observed objects. The drill is securely mounted to prevent safety hazards, and event data is captured simultaneously with readings from an optical tachometer. To ensure sufficient data for analysis, we record for four seconds, guaranteeing at least 80 object revolutions at the drill's lowest speed. Since the drill's speed isn't perfectly constant, we compare the data from both methods for each individual second when evaluating different Region of Interest sizes (\eg Table~\ref{tab:roisizes_line}), as we assume minimal speed variations within each one-second interval.

\subsubsection{Felt disc with a high-contrast mark}
\label{experiment:felt-disc}

\begin{table*}
\begin{tabular}{p{0.31\textwidth} p{0.31\textwidth} p{0.31\textwidth}}
\begin{minipage}{\linewidth}

    \centering
    \scriptsize
    \setlength\tabcolsep{0pt}
    \newcolumntype{C}{>{\centering\arraybackslash}p{5.25cm}}
    \begin{tabular}{C}
    \settoheight{\imageheight}{\includegraphics{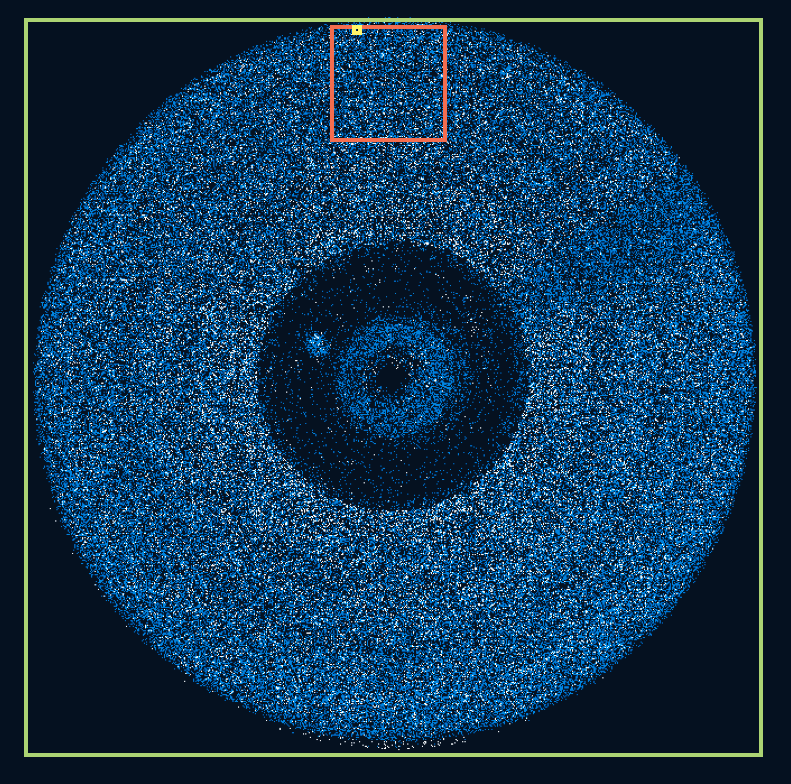}}
    \includegraphics[trim=0 0.7\imageheight{} 0 0, clip, width=0.985\columnwidth]{line_rois_findsize+}\\
    (a) Selected Regions of Interest\\
    \end{tabular}

    \centering
    \scriptsize
    \setlength\tabcolsep{1pt}
    \newcolumntype{C}{>{\centering\arraybackslash}p{1.7cm}}
    \begin{tabular}{CCC}
    \includegraphics[width=0.32\columnwidth]{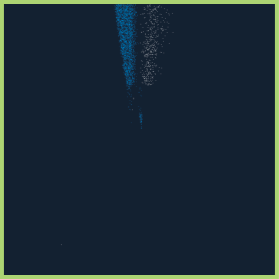} & \includegraphics[width=0.32\columnwidth]{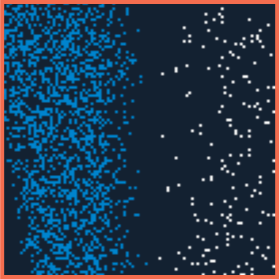} & \includegraphics[width=0.32\columnwidth]{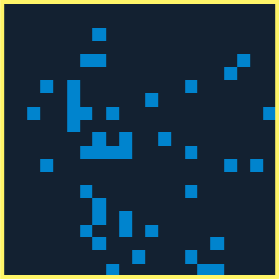}\\
    $655\x655$px & $100\x100$px & $20\x20$px\\
    \multicolumn{3}{c}{\begin{tabular}[c]{@{}c@{}}(b) Templates with aggreg. duration of 0.1 ms.\end{tabular}}\\
    \end{tabular}

    \centering
    \scriptsize
    \setlength\tabcolsep{1.3pt}
    \begin{tabular}{cccc}
    \includegraphics[width=0.23\columnwidth]{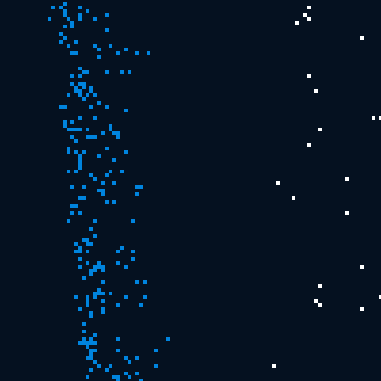} &  \includegraphics[width=0.23\columnwidth]{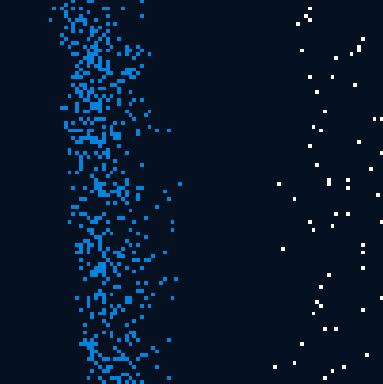} & \includegraphics[width=0.23\columnwidth]{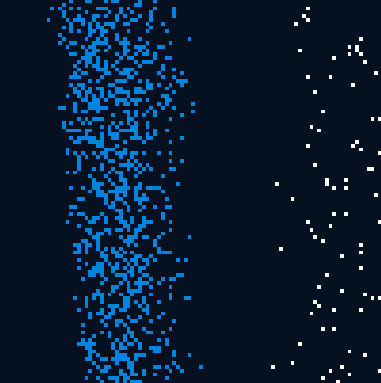} &  \includegraphics[width=0.23\columnwidth]{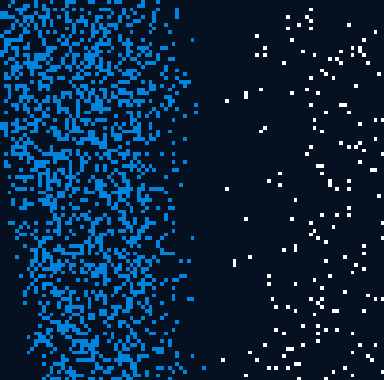}\\
    $t\eq0.1$ ms & $t\eq0.25$ ms & $t\eq0.5$ & $t\eq1$ ms\\
    \multicolumn{4}{c}{\begin{tabular}[c]{@{}c@{}}(c) A template as a function of\end{tabular}}\\
    \multicolumn{4}{c}{\begin{tabular}[c]{@{}c@{}}the duration $t$ of the aggregation time interval.\end{tabular}}
    \end{tabular}

\captionof{figure}{Setup of experiment \ref{experiment:felt-disc} (see Tab.~\ref{tab:roisizes_line}, \ref{tab:windowdur_line}).}
\label{fig:line_pics}

\end{minipage}
&
\begin{minipage}{\linewidth}

    \vspace{-1.65em}
    \centering
    \tiny
    \renewcommand{\arraystretch}{2.26}
    \setlength\tabcolsep{0pt}
    \newcolumntype{C}{>{\centering\arraybackslash}p{1cm}}
    \begin{tabular}{|C||C|C|C|C|}
        \hline
    \backslashbox{\,t(s)}{method\,}& Tacho & 
    \multicolumn{1}{c|}{\begin{tabular}[c]{@{}c@{}}Our method\\ $655\x655$px\end{tabular}} & 
    \multicolumn{1}{c|}{\begin{tabular}[c]{@{}c@{}}Our method\\ $100\x100$px\end{tabular}} & 
    \begin{tabular}[c]{@{}c@{}}Our method\\ $20\x20$px\end{tabular} \\ \hline\hline
         $\lbrack0,1)$ & \begin{tabular}[c]{@{}c@{}}1199.43\\$\pm$ 0.82\end{tabular} & \begin{tabular}[c|]{@{}c@{}}1199.12\\$\pm$ 0.53\end{tabular} & 
                    \begin{tabular}[c]{@{}c@{}}1199.12\\$\pm$ 0.53\end{tabular} & \begin{tabular}[c]{@{}c@{}}1199.12\\$\pm$ 0.53\end{tabular}\\ \hline
                    
         $\lbrack1,2)$ & \begin{tabular}[c]{@{}c@{}}1200.85\\$\pm$ 0.21\end{tabular} & \begin{tabular}[c]{@{}c@{}}1200.48\\$\pm$ 0.64\end{tabular} & 
                    \begin{tabular}[c]{@{}c@{}}1200.48\\$\pm$ 0.73\end{tabular} & \begin{tabular}[c]{@{}c@{}}1200.36\\$\pm$ 0.61\end{tabular}\\ \hline
                    
         $\lbrack2,3)$ & \begin{tabular}[c]{@{}c@{}}1203.18\\$\pm$ 0.41\end{tabular} & \begin{tabular}[c]{@{}c@{}}1202.53\\$\pm$ 0.93\end{tabular} &
                    \begin{tabular}[c]{@{}c@{}}1202.53\\$\pm$ 0.93\end{tabular} & \begin{tabular}[c]{@{}c@{}}1202.65\\$\pm$ 1.07\end{tabular}\\ \hline
                    
         $\lbrack3,4)$ & \begin{tabular}[c]{@{}c@{}}1203.1\\$\pm$ 0.21\end{tabular} & \begin{tabular}[c]{@{}c@{}}1203.49\\$\pm$ 0.87\end{tabular} & 
                    \begin{tabular}[c]{@{}c@{}}1203.49\\$\pm$ 0.72\end{tabular} & \begin{tabular}[c]{@{}c@{}}1203.49\\$\pm$ 0.87\end{tabular}\\ \hline
                    
    \end{tabular}
    \captionof{table}{Revolutions per minute $\pm$ 2$\sigma$~(\ref{eq:sigma}) as a function of the size of the Region of Interest (see Fig.\ref{fig:line_pics}b).}
    \label{tab:roisizes_line}

\end{minipage}
&
\begin{minipage}{\linewidth}

    \vspace{-1.65em}
    \centering
    \tiny
    \renewcommand{\arraystretch}{2.28}
    \setlength\tabcolsep{0pt}
    \newcolumntype{C}{>{\centering\arraybackslash}p{1.7cm}}
    \begin{tabular}{|C||C|C|}
        \hline  
        {\backslashbox{t(ms)}{method}} & 
        Tachometer & 
        \multicolumn{1}{c|}{\begin{tabular}[c]{@{}c@{}}Our method\\$100\x100$px\end{tabular}}\\ \hline\hline
        0.1 & \multirow{4}{*}{\begin{tabular}[c]{@{}c@{}}1198.9\\ $\pm$ 1\end{tabular}} & \begin{tabular}[c]{@{}c@{}}1199.12\\$\pm$0.53\end{tabular} \\ \cline{1-1} \cline{3-3} 
        0.25 &  & \begin{tabular}[c]{@{}c@{}}1199.06\\$\pm$ 1\end{tabular} \\ \cline{1-1} \cline{3-3} 
        0.5 &  & \begin{tabular}[c]{@{}c@{}}1198.75\\$\pm$ 1.67\end{tabular} \\ \cline{1-1} \cline{3-3} 
        1.0 &  & \begin{tabular}[c]{@{}c@{}}1198.76\\$\pm$ 2.41\end{tabular} \\ \hline
    \end{tabular}
    \captionof{table}{Revolutions per minute $\pm$ 2$\sigma$~(\ref{eq:sigma}) as a function of the aggregation duration (see Fig.~\ref{fig:line_pics}c).}
    \label{tab:windowdur_line}

\end{minipage}
\end{tabular}
\vspace{-0.5em}
\captionof{figure}{The fronto-parallel felt disc with a high-contrast mark experiment (see Fig.~\ref{fig:physical_setups}c).}
\vspace{-1em}
\label{fig:line_results}
\end{table*}

This subsection presents experiments with the high-contrast mark setup (see Fig.~\ref{fig:physical_setups}c).

\vspace{-1em}\paragraph{Selection of RoI}
We experimented with different RoI positions and sizes to find the smallest RoI size that still produced accurate results. We picked three of them for demonstration purposes (see Fig.~\ref{fig:line_pics}a,b), with the duration of the event aggregation fixed and set to $0.1$ millisecond. 
The results are presented in Tab.~\ref{tab:roisizes_line} alongside measurements obtained from the laser tachometer. 

The results show that even the smallest RoI of size $20$ by $20$ pixels (px) produces errors of the same order as the larger RoI sizes. 

\vspace{-1em}\paragraph{Selection of aggregation duration}
In this experiment, with the duration of the event aggregation, we fixed the RoI to size $100\times100$ px. We present results from one second of data with various durations of aggregation ranging from $0.1$ millisecond (ms) to $1$ ms. For the templates for each duration of the aggregation, see Fig.~\ref{fig:line_pics}c. The results are shown in Tab.~\ref{tab:windowdur_line}. 
Our method achieved consistent accuracy across all tested aggregation durations, closely matching the ground truth provided by the tachometer. While a slight increase in standard deviation was observed with longer durations, the overall deviation remained minimal. This suggests that the distinct pattern generated by the rotating mark allows for accurate estimation even when aggregating events over a longer time window.

\vspace{-1em}
\subsubsection{Fronto-parallel velcro disc}
\label{experiment:velcro-disc-front}

\begin{table*}
\begin{tabular}{p{0.31\textwidth} p{0.31\textwidth} p{0.31\textwidth}}
\begin{minipage}{\linewidth}

    \centering
    \scriptsize
    \setlength\tabcolsep{0pt}
    \newcolumntype{C}{>{\centering\arraybackslash}p{5.25cm}}
    \begin{tabular}{C}
    \settoheight{\imageheight}{\includegraphics{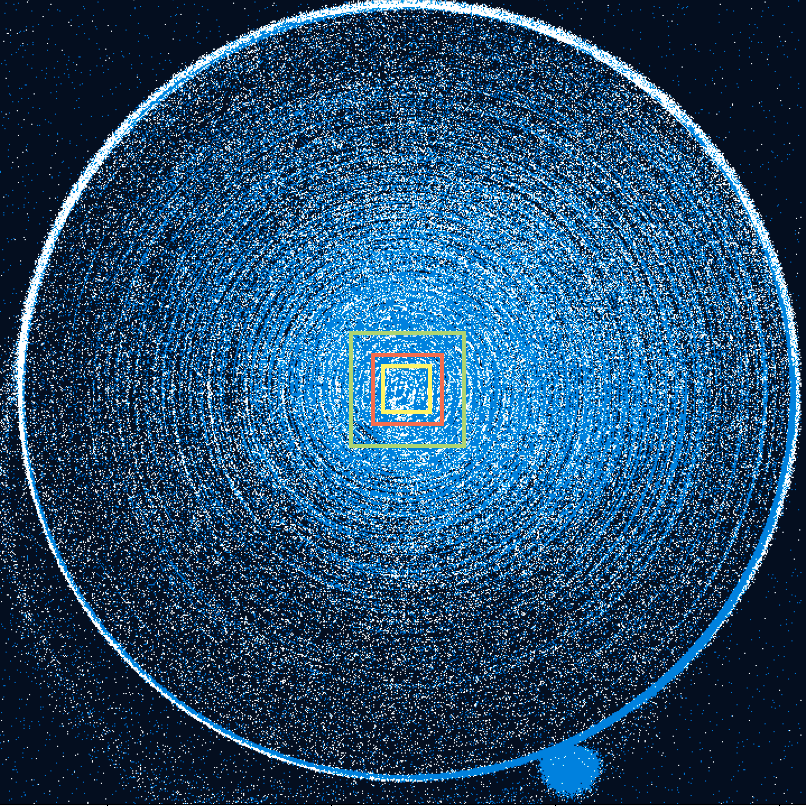}}
    \includegraphics[trim=0 0.35\imageheight{} 0 0.35\imageheight{}, clip, width=0.985\columnwidth]{velcro_rois_findsize+}\\
    (a) Selected Regions of Interest\\
    \end{tabular}

    \centering
    \scriptsize
    \setlength\tabcolsep{1pt}
    \newcolumntype{C}{>{\centering\arraybackslash}p{1.7cm}}
    \begin{tabular}{CCC}
    \includegraphics[width=0.32\columnwidth]{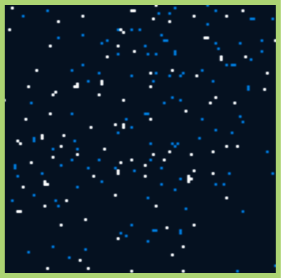} & 
    \includegraphics[width=0.32\columnwidth]{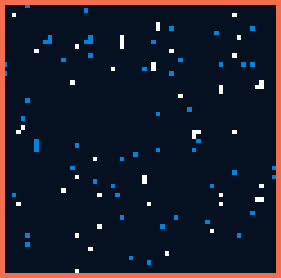} & 
    \includegraphics[width=0.32\columnwidth]{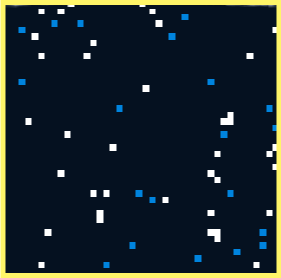}\\
    $100\x100$ px & $60\x60$ px & $40\x40$ px\\
    \multicolumn{3}{c}{\begin{tabular}[c]{@{}c@{}}(b) Templates with aggreg. duration of 0.1 ms.\end{tabular}}\\
    \end{tabular}

    \centering
    \scriptsize
    \setlength\tabcolsep{1.3pt}
    \begin{tabular}{cccc}
    \includegraphics[width=0.23\columnwidth]{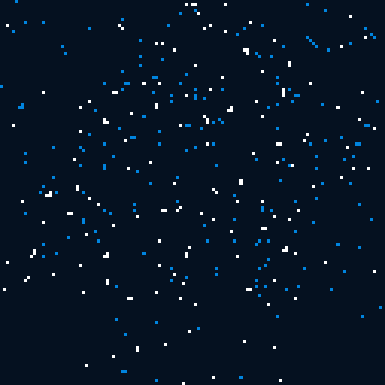} &  
    \includegraphics[width=0.23\columnwidth]{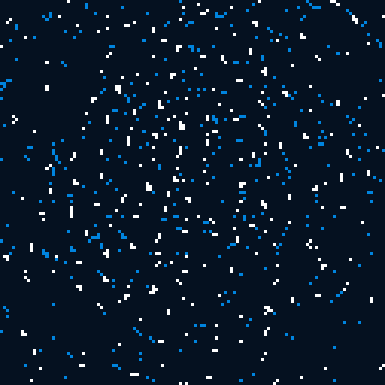} & 
    \includegraphics[width=0.23\columnwidth]{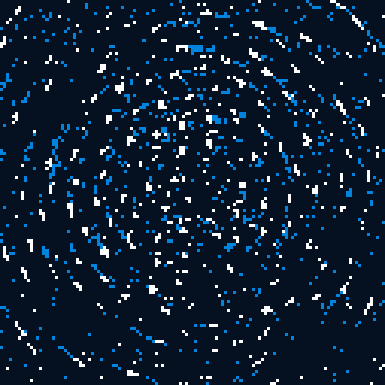} &  
    \includegraphics[width=0.23\columnwidth]{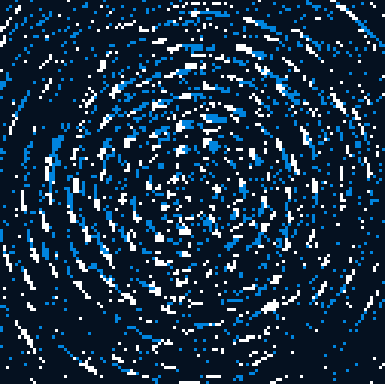}\\
    $t\eq0.1$ ms & $t\eq0.25$ ms & $t\eq0.5$ & $t\eq1$ ms\\
    \multicolumn{4}{c}{\begin{tabular}[c]{@{}c@{}}(c) A template as a function of\end{tabular}}\\
    \multicolumn{4}{c}{\begin{tabular}[c]{@{}c@{}}the duration $t$ of the aggregation time interval.\end{tabular}}
    \end{tabular}

\captionof{figure}{Setup of experiment \ref{experiment:velcro-disc-front} (see Tab.~\ref{tab:roisizes_velcro}, \ref{tab:windowdur_velcro}).}
\label{fig:velcro_pics}

\end{minipage}
&
\begin{minipage}{\linewidth}

    \vspace{-1.65em}
    \centering
    \tiny
    \renewcommand{\arraystretch}{2.26}
    \setlength\tabcolsep{0pt}
    \newcolumntype{C}{>{\centering\arraybackslash}p{1cm}}
    \begin{tabular}{|C||C|C|C|C|}
        \hline
    \backslashbox{\,t(s)}{method\,}& Tacho & 
    \multicolumn{1}{c|}{\begin{tabular}[c]{@{}c@{}}Our method\\ $100\x100$px\end{tabular}} & 
    \multicolumn{1}{c|}{\begin{tabular}[c]{@{}c@{}}Our method\\ $60\x60$px\end{tabular}} & 
    \begin{tabular}[c]{@{}c@{}}Our method\\ $40\x40$px\end{tabular} \\ \hline\hline
         $\lbrack0,1)$ & \begin{tabular}[c]{@{}c@{}}1266.07\\$\pm$ 0.38\end{tabular} & \begin{tabular}[c|]{@{}c@{}}1266.21\\$\pm$ 0.83\end{tabular} & 
                \begin{tabular}[c]{@{}c@{}}1265.96\\$\pm$ 1.36\end{tabular} & \begin{tabular}[c]{@{}c@{}}1394.74\\$\pm$ 269.44\end{tabular}\\ \hline
                
    $\lbrack1,2)$ & \begin{tabular}[c]{@{}c@{}}1266.67\\$\pm$ 0.29\end{tabular} & \begin{tabular}[c]{@{}c@{}}1266.46\\$\pm$ 1.13\end{tabular} & 
                \begin{tabular}[c]{@{}c@{}}1266.72\\$\pm$ 1.41\end{tabular} & \begin{tabular}[c]{@{}c@{}}1670.26\\$\pm$ 365.62\end{tabular}\\ \hline
                
    $\lbrack2,3)$ & \begin{tabular}[c]{@{}c@{}}1267.65\\$\pm$ 0.46\end{tabular} & \begin{tabular}[c]{@{}c@{}}1267.61\\$\pm$ 0.75\end{tabular} & 
                \begin{tabular}[c]{@{}c@{}}1267.61\\$\pm$ 1.04\end{tabular} & \begin{tabular}[c]{@{}c@{}}1235.53\\$\pm$ 61.58\end{tabular}\\ \hline
                
    $\lbrack3,4)$ & \begin{tabular}[c]{@{}c@{}}1267.38\\$\pm$ 0.4\end{tabular} & \begin{tabular}[c]{@{}c@{}}1267.48\\$\pm$ 1.05\end{tabular} & 
                \begin{tabular}[c]{@{}c@{}}1267.36\\$\pm$ 1.6\end{tabular} & \begin{tabular}[c]{@{}c@{}}1407.24\\$\pm$ 219.57\end{tabular}\\ \hline
                
    \end{tabular}
    \captionof{table}{Revolutions per minute $\pm$ 2$\sigma$~(\ref{eq:sigma}) as a function of the size of the Region of Interest (see Fig.\ref{fig:velcro_pics}b).}
    \label{tab:roisizes_velcro}

\end{minipage}
&
\begin{minipage}{\linewidth}

    \vspace{-1.65em}
    \centering
    \tiny
    \renewcommand{\arraystretch}{2.28}
    \setlength\tabcolsep{0pt}
    \newcolumntype{C}{>{\centering\arraybackslash}p{1.7cm}}
    \begin{tabular}{|C||C|C|}
        \hline  
        {\backslashbox{\,t(ms)}{method\,}}&Tachometer& 
        \multicolumn{1}{c|}{\begin{tabular}[c]{@{}c@{}}Our method\\$120\x120$px\end{tabular}}\\ \hline\hline
        0.1 & \multirow{4}{*}{\begin{tabular}[c]{@{}c@{}}1266.08\\ $\pm$ 0.38\end{tabular}} & \begin{tabular}[c]{@{}c@{}}1266.08\\$\pm$0.71\end{tabular} \\ \cline{1-1} \cline{3-3} 
        0.25 &  & \begin{tabular}[c]{@{}c@{}}1265.83\\$\pm$ 1.46\end{tabular} \\ \cline{1-1} \cline{3-3} 
        0.5 &  & \begin{tabular}[c]{@{}c@{}}1265.85\\$\pm$ 2.4\end{tabular} \\ \cline{1-1} \cline{3-3} 
        1.0 &  & \begin{tabular}[c]{@{}c@{}}1265.96\\$\pm$ 5.83\end{tabular} \\ \hline
    \end{tabular}
    \captionof{table}{Revolutions per minute $\pm$ 2$\sigma$~(\ref{eq:sigma}) as a function of the aggregation duration (see Fig.~\ref{fig:velcro_pics}c).}
    \label{tab:windowdur_velcro}

\end{minipage}
\end{tabular}
\vspace{-0.5em}
\captionof{figure}{The fronto-parallel velcro disc experiment (see Fig.~\ref{fig:physical_setups}d).}
\vspace{0.5em}
\label{fig:velcro_results}
\end{table*}

Here, we present experiments in the velcro disc setup (see Fig.~\ref{fig:physical_setups}d) with fronto-parallel camera position.

\vspace{-1em}\paragraph{Selection of RoI} For the demonstration purposes, we picked three Regions of Interest of sizes $100\times100$ px, $60\times60$ px and $40\times40$ px (see Fig.~\ref{fig:velcro_pics}a,b). As shown in Tab.~\ref{tab:roisizes_velcro}, when a distinguishable pattern is not present in the template, the smallest RoI of size $40\times40$ px produces errors of two orders larger than in the case of larger Regions of Interest.

\vspace{-1em}\paragraph{Selection of aggregation duration}
We fixed the RoI size to $120\times120$ px and aligned it with the object's centre of rotation. From Tab.~\ref{tab:windowdur_velcro}, we see that the average RPM values remain close to those measured by the tachometer. The standard deviation of the average RPM increases as the duration of the event aggregation prolongs, which is expected as longer time intervals of event aggregation reduce accuracy. 

\vspace{-1em}
\subsubsection{Velcro disc with non-frontal camera behind a glass sheet}
\label{experiment:velcro-disc-side}

\begin{table*}
\begin{tabular}{p{0.31\textwidth} p{0.31\textwidth} p{0.31\textwidth}}
\begin{minipage}{\linewidth}

    \centering
    \scriptsize
    \setlength\tabcolsep{0pt}
    \newcolumntype{C}{>{\centering\arraybackslash}p{5.25cm}}
    \begin{tabular}{C}
    \settoheight{\imageheight}{\includegraphics{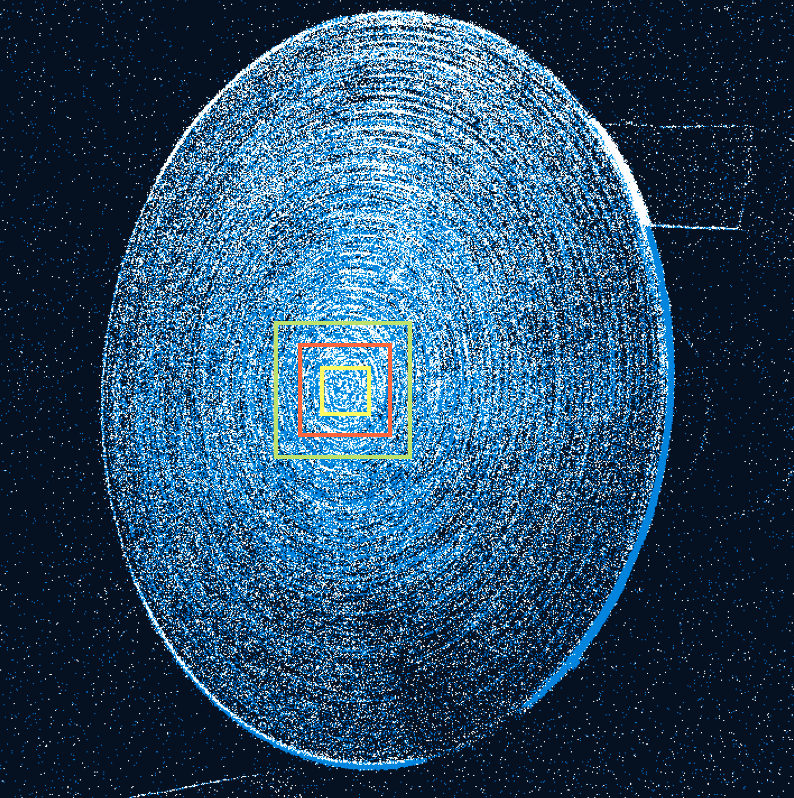}}
    \includegraphics[trim=0 0.35\imageheight{} 0 0.35\imageheight{}, clip, width=0.985\columnwidth]{velcroside_rois_findsize+}\\
    (a) Selected Regions of Interest\\
    \end{tabular}

    \centering
    \scriptsize
    \setlength\tabcolsep{1pt}
    \newcolumntype{C}{>{\centering\arraybackslash}p{1.7cm}}
    \begin{tabular}{CCC}
    \includegraphics[width=0.32\columnwidth]{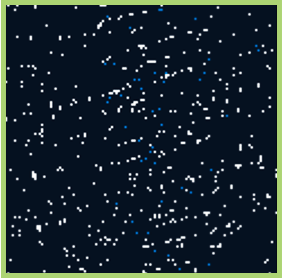} & \includegraphics[width=0.32\columnwidth]{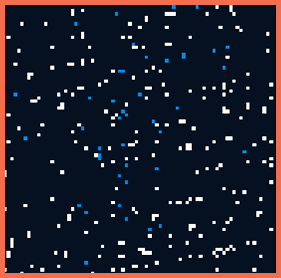} & \includegraphics[width=0.32\columnwidth]{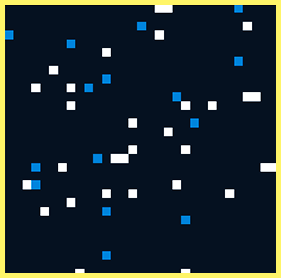}\\
    $120\x120$ px & $80\x80$ px & $35\x35$ px\\
    \multicolumn{3}{c}{\begin{tabular}[c]{@{}c@{}}(b) Templates with aggreg. duration of 0.1 ms.\end{tabular}}\\
    \end{tabular}

    \centering
    \scriptsize
    \setlength\tabcolsep{1.3pt}
    \begin{tabular}{cccc}
    \includegraphics[width=0.23\columnwidth]{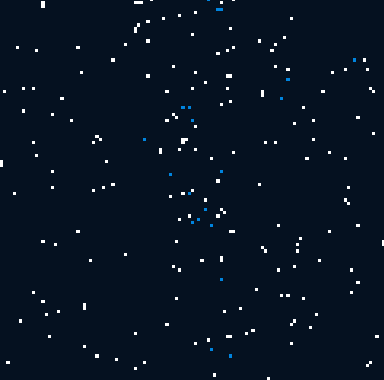} &  \includegraphics[width=0.23\columnwidth]{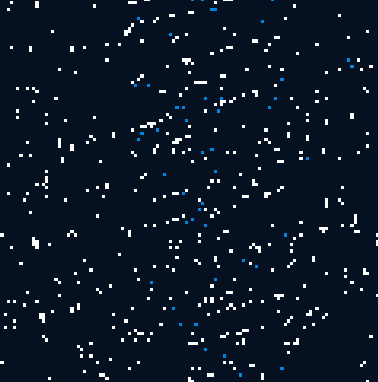} & \includegraphics[width=0.23\columnwidth]{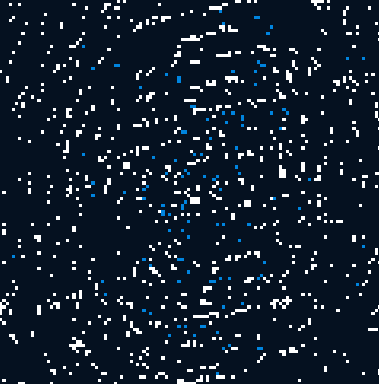} &  \includegraphics[width=0.23\columnwidth]{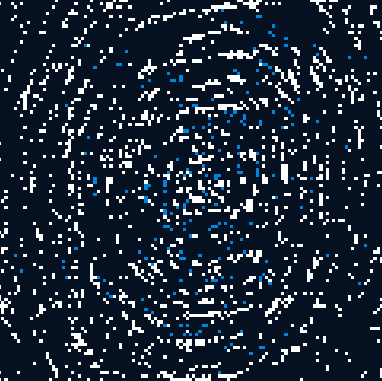}\\
    $t\eq0.1$ ms & $t\eq0.25$ ms & $t\eq0.5$ & $t\eq1$ ms\\
    \multicolumn{4}{c}{\begin{tabular}[c]{@{}c@{}}(c) A template as a function of\end{tabular}}\\
    \multicolumn{4}{c}{\begin{tabular}[c]{@{}c@{}}the duration $t$ of the aggregation time interval.\end{tabular}}
    \end{tabular}

\captionof{figure}{Setup of experiment \ref{experiment:velcro-disc-side} (see Tab.~\ref{tab:roisizes_velcroside}, \ref{tab:windowdur_velcroside}).}
\label{fig:velcroside_pics}

\end{minipage}
&
\begin{minipage}{\linewidth}

    \vspace{-0.4em}
    \centering
    \tiny
    \renewcommand{\arraystretch}{2.26}
    \setlength\tabcolsep{0pt}
    \newcolumntype{C}{>{\centering\arraybackslash}p{1cm}}
    \begin{tabular}{|C||C|C|C|C|}
        \hline
    \backslashbox{\,t(s)}{method\,}& Tacho & 
    \multicolumn{1}{c|}{\begin{tabular}[c]{@{}c@{}}Our method\\ $120\x120$px\end{tabular}} & 
    \multicolumn{1}{c|}{\begin{tabular}[c]{@{}c@{}}Our method\\ $80\x80$px\end{tabular}} & 
    \begin{tabular}[c]{@{}c@{}}Our method\\ $35\x35$px\end{tabular} \\ \hline\hline
        $\lbrack0,1)$ & \begin{tabular}[c]{@{}c@{}}1578.2\\$\pm$ 1.48\end{tabular} & \begin{tabular}[c|]{@{}c@{}}1578.56\\$\pm$ 1.78\end{tabular} & 
                    \begin{tabular}[c]{@{}c@{}}1578.56\\$\pm$ 1.78\end{tabular} & \begin{tabular}[c]{@{}c@{}}1257.61\\$\pm$ 192.37\end{tabular}\\ \hline
                    
        $\lbrack1,2)$ & \begin{tabular}[c]{@{}c@{}}1580.19\\$\pm$ 0.98\end{tabular} & \begin{tabular}[c]{@{}c@{}}1579.76\\$\pm$ 1.57\end{tabular} & 
                    \begin{tabular}[c]{@{}c@{}}1579.77\\$\pm$ 2.24\end{tabular} & \begin{tabular}[c]{@{}c@{}}1392.5\\$\pm$ 147.16\end{tabular}\\ \hline
                    
        $\lbrack2,3)$ & \begin{tabular}[c]{@{}c@{}}1578.38\\$\pm$ 0.78\end{tabular} & \begin{tabular}[c]{@{}c@{}}1578.57\\$\pm$ 2.11\end{tabular} & 
                    \begin{tabular}[c]{@{}c@{}}1578.97\\$\pm$ 2.26\end{tabular} & \begin{tabular}[c]{@{}c@{}}1435.02\\$\pm$ 130.11\end{tabular}\\ \hline
                    
        $\lbrack3,4)$ & \begin{tabular}[c]{@{}c@{}}1577.47\\$\pm$ 0.68\end{tabular} & \begin{tabular}[c]{@{}c@{}}1577.76\\$\pm$ 1.29\end{tabular} & 
                    \begin{tabular}[c]{@{}c@{}}1577.36\\$\pm$ 1.85\end{tabular} & \begin{tabular}[c]{@{}c@{}}1394\\$\pm$ 166.97\end{tabular}\\ \hline 
    \end{tabular}
    \captionof{table}{Revolutions per minute $\pm$ 2$\sigma$~(\ref{eq:sigma}) as a function of the size of the Region of Interest (see Fig.\ref{fig:velcroside_pics}b).}
    \label{tab:roisizes_velcroside}

\end{minipage}
&
\begin{minipage}{\linewidth}

    \vspace{-1.65em}
    \centering
    \tiny
    \renewcommand{\arraystretch}{2.28}
    \setlength\tabcolsep{0pt}
    \newcolumntype{C}{>{\centering\arraybackslash}p{1.7cm}}
    \begin{tabular}{|C||C|C|}
        \hline  
        {\backslashbox{\,t(ms)}{method\,}}&Tachometer& 
        \multicolumn{1}{c|}{\begin{tabular}[c]{@{}c@{}}Our method\\$120\x120$px\end{tabular}}\\ \hline\hline
        0.1 & \multirow{4}{*}{\begin{tabular}[c]{@{}c@{}}1578.2\\ $\pm$ 1.48\end{tabular}} & \begin{tabular}[c]{@{}c@{}}1578.29\\$\pm$ 1.21\end{tabular} \\ \cline{1-1} \cline{3-3} 
        0.25 &  & \begin{tabular}[c]{@{}c@{}}1578.56\\$\pm$ 1.78\end{tabular} \\ \cline{1-1} \cline{3-3} 
        0.5 &  & \begin{tabular}[c]{@{}c@{}}1578.16\\$\pm$ 1.55\end{tabular} \\ \cline{1-1} \cline{3-3} 
        1.0 &  & \begin{tabular}[c]{@{}c@{}}1578.95\\$\pm$ 3.05\end{tabular} \\ \hline
    \end{tabular}
    \captionof{table}{Revolutions per minute $\pm$ 2$\sigma$~(\ref{eq:sigma}) as a function of the aggregation duration (see Fig.~\ref{fig:velcroside_pics}c).}
    \label{tab:windowdur_velcroside}

\end{minipage}
\end{tabular}
\vspace{-0.5em}
\captionof{figure}{The non-frontal velcro disc experiment (see Fig.~\ref{fig:physical_setups}e).}
\vspace{-1em}
\label{fig:velcroside_results}
\end{table*}

In this subsection, we present experiments with a velcro disc observed by the camera at a 45\textdegree~angle that captures data through a sheet of glass (see~Fig.\ref{fig:physical_setups}e). 

\vspace{-1em}\paragraph{Selection of RoI} We experiment with RoI sizes ranging from $200 \times 200$ px to $35 \times 35$ px. For RoI positions, see Fig.~\ref{fig:velcroside_pics}a. We present results for three selected sizes in Tab.~\ref{tab:roisizes_velcroside}. As shown in this table, the performance of our method degrades significantly in the case of the smallest $35 \times 35$ px RoI. We believe that it is caused by the fact that there are not enough distinctive events in such a small RoI.

\vspace{-1em}\paragraph{Selection of aggregation duration} From Tab.~\ref{tab:windowdur_velcroside}, it is clear that with $120 \times 120$ px RoI, all aggregation durations yield measurements, closely reflecting the ground truth provided by the laser tachometer. 
While an increase in measurement uncertainty ($2\sigma$) was observed with the longest duration, it remained relatively small. 
This suggests that our method can effectively estimate the rotational speed with minimal impact from aggregation duration within the tested range, even for objects with less distinct patterns, like the uniform velcro surface.

\subsection{Discussion}

Based on the presented experiments, we conclude that 
(i) a small RoI is feasible without degraded accuracy when a distinguishable pattern is present. 
(ii) the best results are achieved when the RoI covers the area with the highest density of events, and the template captures a distinctive pattern emerging periodically, 
(iii) the event aggregation duration of $0.25$ ms is preferred, as it provides a good balance between a low number of event-aggregation arrays resulting in faster computations and relatively low standard deviation of the average results.

\vspace{-1em}\paragraph{Limitations} The presented method does not consider an automatic detection of a suitable RoI and its respective template. Also, 
no centrosymmetric objects were tested - the symmetries might produce spurious peaks. With non-stationary scenes, these limitations remain the main topic for future works.

\section{Conclusion}

In this paper, we proposed a novel contactless measurement method of periodic phenomena with an event camera. The method only assumes that the observed object periodically produces a similar set of events by returning to a known state or position.

We evaluated the proposed method on the task of measuring the frequency of periodic phenomena and rotational speed, achieving a relative error lower than $\pm0.04\%$, which is within the error margin of the ground-truth measurement. 
The precision is maintained while measuring frequencies ranging from 20 hertz (equivalent to 1200 RPM) up to 2 kilohertz (equivalent to 120 000 RPM). We demonstrated robustness against changes in camera angles.

Our dataset of captured data is publicly available on GitHub at \href{https://github.com/JackPieCZ/EE3P}{JackPieCZ/EE3P}.

\scriptsize\paragraph{Acknowledgement} The authors acknowledge the Grant Agency of the Czech Technical University in Prague, grant No.SGS23/173/OHK3/3T/13.

\newpage
{\small
\bibliographystyle{hieeetr}
\bibliography{cvww_paper}
}

\end{document}